\PassOptionsToPackage{table}{xcolor}
\documentclass{article}
\usepackage{iclr2027_conference,times}
\usepackage[T1]{fontenc}
\usepackage[type1,semibold]{plex-sans}
\usepackage{amsmath,amssymb,booktabs,graphicx,hyperref,url,longtable,listings}
\usepackage[table]{xcolor}
\usepackage{tabularx}
\definecolor{WWBBlue}{HTML}{5271AE}
\colorlet{WWBHeader}{WWBBlue!13}
\colorlet{WWBStripe}{WWBBlue!4}
\newcolumntype{L}{>{\raggedright\arraybackslash}X}
\newcolumntype{R}{>{\raggedleft\arraybackslash}X}
\hypersetup{hidelinks,pdftitle={WhatWorkedBench: Benchmarking Experimental Understanding in AI Agents},pdfauthor={Jingjie Ning, Xueqi Li, Yibo Kong, Dongting Li}}
\newcommand{\CatalogSources}{30}
\newcommand{\CatalogFamilies}{8}
\newcommand{\CatalogConditions}{36}
\newcommand{\CatalogConfigurations}{1248}
\newcommand{\CatalogScoreChecks}{4206}
\newcommand{\CatalogReplaySeconds}{127.87}

\newcommand{\AgentTotalEpisodes}{108}
\newcommand{\AgentTotalValid}{94}
\newcommand{\AdditionalInformationValid}{19}

\title{WhatWorkedBench: Benchmarking\\Experimental Understanding in AI Agents}
\iclrfinalcopy
\author{%
\makebox[\dimexpr\textwidth-2\tabcolsep\relax][c]{%
\begin{tabular}{c}
Jingjie Ning\textsuperscript{1}\thanks{Corresponding author}\quad
Xueqi Li\textsuperscript{1}\quad
Yibo Kong\textsuperscript{1}\quad
Dongting Li\textsuperscript{2}\\[4pt]
{\normalfont\small \textsuperscript{1}Carnegie Mellon University\quad
\textsuperscript{2}Tsinghua University}\\[2pt]
{\normalfont\footnotesize
\{\href{mailto:jening@cs.cmu.edu}{jening},\,
\href{mailto:xueqil@cs.cmu.edu}{xueqil},\,
\href{mailto:yibok@cs.cmu.edu}{yibok}\}@cs.cmu.edu\quad
\href{mailto:ldt25@mails.tsinghua.edu.cn}{ldt25@mails.tsinghua.edu.cn}}
\end{tabular}}%
}

\begin{document}
\maketitle
\lhead{Preprint}
\vspace{-.25in}
\noindent\makebox[\textwidth][c]{\href{https://ethanning.github.io/WhatWorkedBench/}{\includegraphics[height=.25in]{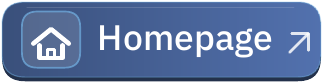}}\hspace{.7em}\href{https://github.com/EthanNing/WhatWorkedBench}{\includegraphics[height=.25in]{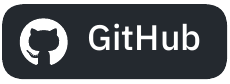}}}
\par\vspace{.04em}
\AddToHook{env/tabular/begin}{\sffamily}
\AddToHook{env/tabular*/begin}{\sffamily}
\AddToHook{env/longtable/begin}{\sffamily}
\AddToHook{env/tabularx/begin}{\sffamily\rowcolors{2}{white}{WWBStripe}\renewcommand{\arraystretch}{1.12}\setlength{\aboverulesep}{0pt}\setlength{\belowrulesep}{0pt}}

\begin{abstract}
AI research agents must predict the effects of computational changes after
budgeted experiments. WhatWorkedBench evaluates this \emph{experimental
understanding} through a delivered \emph{response surface} of configuration
scores. Agents inspect workflow code and buy measurements; exhaustive CPU
references score conditional component effects, mean pair interactions,
configuration choice, and delivery. The catalog spans \CatalogConditions{}
task conditions, \CatalogSources{} sources, \CatalogFamilies{} workflow
families, and \CatalogConfigurations{} indexed configuration records, with 4,206 numerical
controls. At eight purchased measurements plus two free anchors, pair-effect
ridge selects an exact optimum on 15 of 22 sources; 13 of these cases have
at least one conditional-effect error exceeding 10\% of the task utility
range. Shared-estimator comparisons measure acquisition and reconstruction
on common observations. In a prospective typed study on 12 four-factor
sources, DeepSeek Flash submits 12/12 direct tables and gains 0.147 recovery over a
Gaussian process (GP) fitted to the same observations. Pro delivers 11/12
artifacts, with an all-attempt GP difference of $-0.001$ and a delivered-only
difference of $+0.063$. On six prespecified new agent-evaluation sources,
Flash and Pro gains are 0.149 and 0.074. In eight typed six-factor episodes,
seven final tables satisfy verified code equivalences. Four fresh agents pass
all six registered rules through named estimators and deliver consistent
tables at 0.682 recovery versus 0.710 for separate direct-table runs.
WhatWorkedBench links acquisition, inference, program structure, and delivery.
\end{abstract}

\begin{figure}[!ht]
\centering
\includegraphics[width=\linewidth]{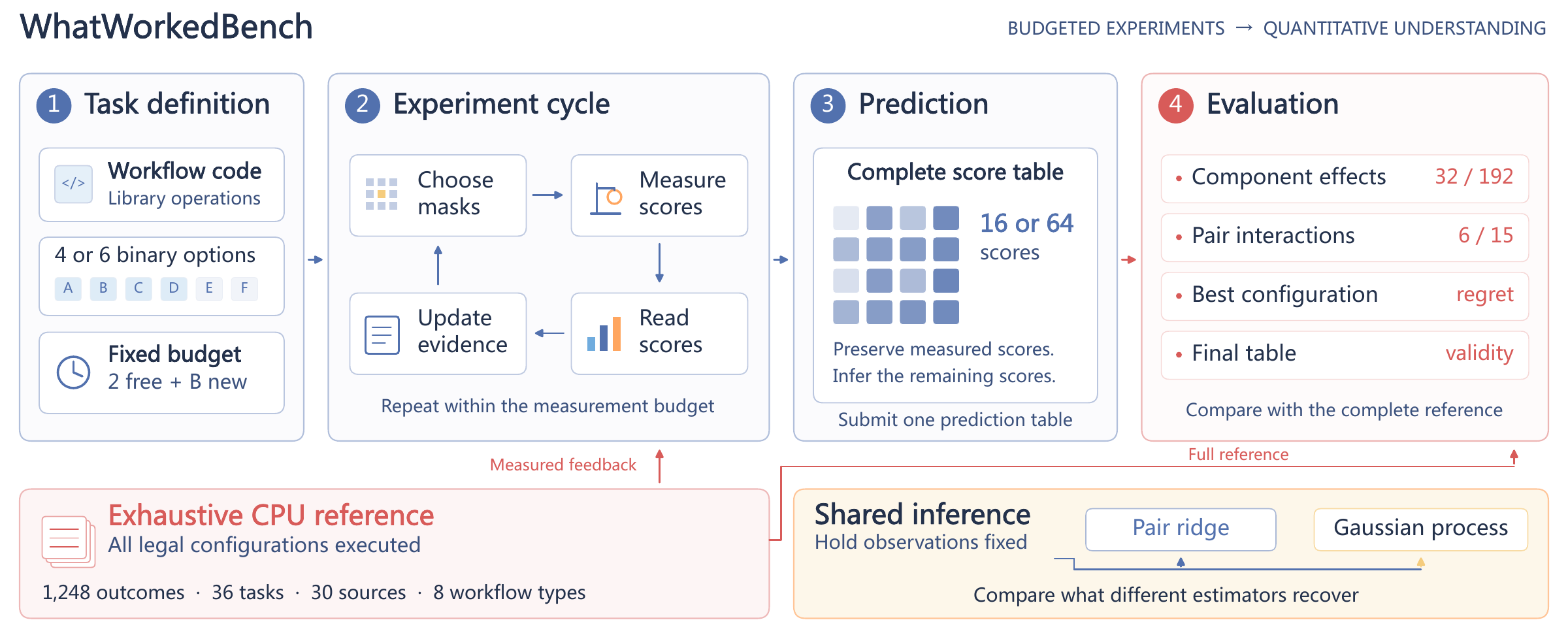}
\caption{WhatWorkedBench connects budgeted experiments to quantitative understanding.
Agents inspect a workflow, query selected configurations, and submit a complete
prediction table. The evaluator uses exhaustive CPU outcomes to score conditional
effects, mean pair interactions (departures from additive effects), configuration
choice, and artifact delivery.
Shared inference compares estimators on the same observations.}
\label{fig:teaser}
\end{figure}

\section{Introduction}
AI agents increasingly plan, execute, and interpret computational
experiments~\citep{mlagentbench,mlebench,expbench}. Research agents use trial
histories to revise training recipes and carry lessons from cheap experiments
to costly configurations~\citep{specialistrecipes,autollmresearch}.
Reliable experimentation requires predicting how component changes behave in
new settings and combine with other changes. Measuring this acquired knowledge
is central to evaluating systems that learn through experimentation.

We study \emph{experimental understanding} as the accuracy of quantitative
predictions about computational interventions available at the end of an
experimental episode. The evaluated system combines an agent's measurement
policy, numerical tools, and submitted prediction artifact. A \emph{native outcome}
is the score of a code-option intervention with source data, implementation,
and evaluation cohort fixed. An agent inspects workflow code and
component definitions, chooses experiments under a measurement budget, and
submits a complete predicted response surface (Figure~\ref{fig:teaser}). A \emph{background} fixes the settings of every other component. The reference
evaluates each component's effect in every background. These conditional
effects describe arbitrary combinations of the documented interventions. They connect the experiments an agent
chooses to the predictive knowledge that its final artifact contains.

\begin{figure}[ht]
\centering
\includegraphics[width=.95\linewidth]{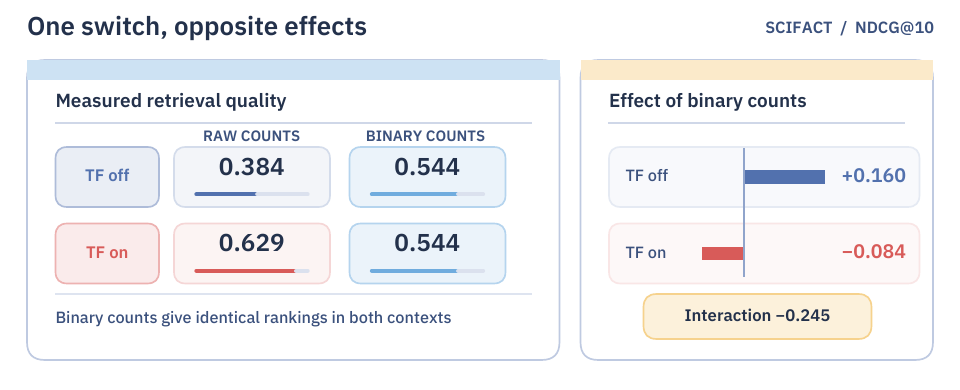}
\caption{A retrieval switch reverses its effect across two workflow backgrounds.
The left panel reports SciFact normalized discounted cumulative gain at rank 10
(NDCG@10) for raw and binary term counts with sublinear term frequency (TF)
off or on. The right panel shows that binary counts raise NDCG@10 by 0.160
with TF off and lower it by 0.084 with TF on, yielding a $-0.245$ interaction.
Binary counts make the sublinear transform inactive, so both right-column
cells have identical rankings and score 0.544. Other settings use unigrams,
English stopword removal, disabled inverse document frequency, and
unnormalized vectors.}
\label{fig:example}
\end{figure}

Figure~\ref{fig:example} illustrates why the full intervention context matters.
The same retrieval option improves one configuration and degrades another.
This dependence of one change on another is an \emph{interaction}.
Across the catalog, 20 of 30 distinct sources, representing 24 of 36 indexed
task conditions, contain a sign reversal with
positive and negative effects each exceeding 5\% of the response range;
At a $10^{-12}$ numerical floor, 35 of 36 conditions and 29 of 30 sources
contain a reversal. A single preferred
configuration or an average component effect leaves
these conditional behaviors unresolved. WhatWorkedBench makes them explicit
quantitative prediction targets.

The benchmark connects three algorithmic decisions. \emph{Measurement selection}
determines which evidence is acquired. \emph{Numerical inference} predicts
unmeasured outcomes from that evidence. \emph{Program structure} supplies
relations among configurations, including inactive options. A
\emph{shared-estimator cross-comparison} fits one estimator to observations
from different acquisition policies and fits different estimators to fixed
observations. Each contrast holds one part of the executable task fixed and
tracks how recovery changes with the other.

\begin{itemize}
\item We define an executable evaluation of experimental understanding through
complete conditional component effects. The reference covers arbitrary
combinations of each task's documented changes, alongside configuration choice
and final artifact delivery.
\item We provide \CatalogConditions{} task conditions on \CatalogSources{}
sources across \CatalogFamilies{} computational workflow families. Exhaustive
native execution yields \CatalogConfigurations{} indexed configuration records
(1,152 distinct source--configuration outcomes), 3,392 indexed conditional
effects (including 192 paired-subcube repeats), and 342 mean pair interactions,
with standalone execution and grading.
\item Across 4,206 numerical controls, \AgentTotalEpisodes{} core agent episodes,
and 24 typed episodes on 12 sources, we measure acquisition, numerical
inference, program structure, and delivery. Across all 12 typed sources,
Flash gains 0.147 recovery over the same-observation GP; Pro has an
all-attempt difference of $-0.001$ and a delivered-only difference of $+0.063$.
Six new sources yield Flash and Pro gains of 0.149 and 0.074. Separate
diagnostics trace code-derived rules into final tables.
\end{itemize}

\section{Related Work}
\label{sec:related}
\textbf{Interactive discovery.} ScienceWorld and DiscoveryWorld assess scientific
reasoning in interactive environments~\citep{scienceworld,discoveryworld}.
CausalGame evaluates experimental design and causal explanations under hidden
confounding~\citep{causalgame}. Table~\ref{tab:related_targets} maps nearby
benchmarks to their experimental settings and scored artifacts.

\begin{table}[!ht]
\caption{Experimental settings and scored artifacts in nearby evaluations.}
\label{tab:related_targets}
\centering\footnotesize
\setlength{\tabcolsep}{3pt}
\begin{tabularx}{\linewidth}{@{}>{\raggedright\arraybackslash}p{3.2cm}L@{}}
\toprule
\rowcolor{WWBHeader}
\textbf{Benchmark} & \textbf{Setting and scored artifact} \\
\midrule
BoxingGym~\citeyearpar{boxinggym} & Scientific worlds; experiment choice and prediction \\
Gravity-Bench~\citeyearpar{gravitybench} & Gravity simulations; numerical discovery answers \\
Science-Gym~\citeyearpar{sciencegym};\newline SciGym~\citeyearpar{scigym} & Physics and biology; equations and system reconstruction \\
Science sandboxes\newline\citeyearpar{sciencesandboxes} & Biological loops; optimization and scientific understanding \\
CAFE~\citeyearpar{cafe} & Compound-AI pipelines; factorial effects and interactions \\
petri-bench~\citeyearpar{petribench} & Perturbed simulations; changed factors and interaction signs \\
AstaBench~\citeyearpar{astabench} & Research workflows; artifacts and task outcomes \\
\textbf{WhatWorkedBench} & Budgeted native workflows; complete score tables and conditional effects \\
\bottomrule
\end{tabularx}
\end{table}

\textbf{Research workflows.} AbGen, AblationBench, and SCOPE assess experiment
plans~\citep{abgen,ablationbench,scope}. MLAgentBench, MLE-bench, BLADE,
ScienceAgentBench, EXP-Bench, and ResearchGym evaluate executable research
outcomes~\citep{mlagentbench,mlebench,blade,scienceagentbench,expbench,researchgym}.
FIRE-Bench scores rediscovered findings; AgentActionBench scores reproduction
traces~\citep{firebench,agentactionbench}.

\textbf{Experimental design.} Factorial design and computer experiments study
component effects and response prediction~\citep{box2005statistics,sacks1989computer}.
Hyperparameter optimization studies conditional search and GP-guided
acquisition~\citep{bergstra2011,smac3,snoek2012practical}; fANOVA summarizes
variation over configuration distributions~\citep{fanova}. Our complete
response surfaces score conditional effects and mean pair interactions after
budgeted measurements. Shared-observation comparisons connect this evaluation
to Bayesian experimental design~\citep{chaloner1995bayesian,modernbed}.
Appendix~\ref{app:related} expands the comparison.

\section{Task and evaluation protocol}
For task $t$, native execution defines a deterministic utility
$u_t(z)\in[0,1]$ on configurations $z\in\{0,1\}^{d_t}$, with $d_t\in\{4,6\}$. Each bit is a binary option, called a \emph{factor}; the full vector is a
\emph{mask}. All masks are legal. The agent receives two free anchor measurements,
$u_t(0^{d_t})$ and $u_t(1^{d_t})$, and may purchase at most $B$ additional
distinct measurements. It can query adaptively or in batches. Repeated
measurements are free and over-budget batches are rejected atomically.
Each utility table is defined by the case's frozen algorithm implementation,
random seed, and evaluation cohort.
Four- and six-factor designs permit exhaustive, independently replayed
references while varying how much of the response surface a budget reveals.

The final artifact is a prediction $\hat u_t(z)$ for every legal mask, with
finite values in $[0,1]$. Grading inserts authoritative observations at measured
cells. Submission closes the episode and establishes the delivered artifact
used for the final score. The protocol records intermediate calculations as
part of the experimental trajectory. Measured endpoint pairs give exact effects;
the remaining contrasts require predictions at one or both endpoints.
Appendix~\ref{app:calibration} reports coverage and prediction error.

\paragraph{Primary numerical target.}
For component $i$ and background $z$ with $z_i=0$, define
\begin{equation}
\Delta_i(z)=u_t(z+e_i)-u_t(z),\qquad
E_t=\frac{1}{d_t2^{d_t-1}}\sum_i\sum_{\substack{z\in\{0,1\}^{d_t}\\z_i=0}}
|\widehat\Delta_i(z)-\Delta_i(z)|.
\end{equation}
These effects describe every joint intervention. For configurations $x,y$, a
path that switches their differing bits gives
\begin{equation}
u_t(y)-u_t(x)=\sum_{e\in\operatorname{path}(x,y)}s_e\Delta_e,
\qquad s_e\in\{-1,1\}.
\label{eq:path}
\end{equation}
Thus complete conditional effects and one anchor determine the whole response
surface. Deriving all predicted effects from a single submitted table preserves
this path consistency. If every conditional-effect error is at most $\epsilon$,
the error of any $k$-component intervention is at most $k\epsilon$.
Appendix~\ref{app:metrics} gives the derivation and a configuration-choice
counterexample that clarifies the need for an effect-based evaluation target.

There are 32 conditional effects at four factors and 192 at six. Let
$A_t$ be the mean absolute true conditional effect. We report raw effect MAE
$E_t$ and relative recovery $R_t=\max(0,1-E_t/A_t)$ for $A_t>10^{-12}$.
For a numerically flat table, recovery is one only when $E_t\leq10^{-12}$.
A zero recovery score corresponds to error at least as large as the mean
true-effect magnitude. Reporting $E_t$, $A_t$, and the utility range preserves
the numerical scale of each reconstruction problem.

\paragraph{Additional numerical outputs.}
For every pair of components, we additionally compare the mean second difference
$u(z+e_i+e_j)-u(z+e_i)-u(z+e_j)+u(z)$, averaging over the remaining backgrounds.
These averages summarize pair interactions; the complete conditional effects
retain their dependence on other components. We report MAE over the six or
15 pair contrasts, together with grid MAE and
selection regret $\max_z u_t(z)-u_t(\arg\max_z\hat u_t(z))$. Prediction ties
prefer fewer enabled options and then the lexically smaller mask. Exact optimality allows regret up to $10^{-12}$.

The strict reconstruction score requires every conditional-effect error to be
at most $\max[10^{-12},\tau(\max u_t-\min u_t)]$. We set the development tolerance
to $\tau=0.1$ and evaluate a five-tolerance profile alongside continuous recovery.
Missing delivery receives zero recovery and a strict score of zero. Raw errors include valid artifacts with explicit sample counts.

\section{Native catalog and reference validation}
A \emph{source} is a dataset or record; a \emph{task condition} pairs a source
with a set of workflow options. A \emph{family} groups tasks by computation and
metric. Table~\ref{tab:catalog} summarizes the catalog. The four-factor track contains
22 instances. A paired track adds two options on one source per family,
producing six six-factor variants. The six-factor track additionally includes
four beat-detection records and four graph instances.
Thus there are \CatalogConditions{} task conditions on \CatalogSources{}
sources, comprising 352 four-factor and 896 six-factor configuration records.
Of these, 96 reproduce original subcube outcomes.

\begin{table}[!ht]
\caption{Task conditions and source inventory. Paired variants share their
original sources. Source metadata retain dataset and library relationships,
including the classifier reused for link prediction.}
\vspace{4pt}
\centering\small
\begin{tabularx}{\linewidth}{@{}lrrrL@{}}
\toprule
\rowcolor{WWBHeader}
\textbf{Workflow} & \textbf{$d=4$} & \textbf{$d=6$} & \textbf{Sources} & \textbf{Native operation and metric} \\
\midrule
Classification & 4 & 1 & 4 & Logistic regression; balanced accuracy \\
Regression & 4 & 1 & 4 & Ridge regression; transformed MAE \\
Clustering & 4 & 1 & 4 & $k$-means; normalized mutual information \\
Forecasting & 4 & 1 & 4 & One-step ridge forecasts; transformed MAE \\
Restoration & 4 & 1 & 4 & Native image filters; transformed pixel RMSE \\
Retrieval & 2 & 1 & 2 & TF--IDF retrieval; NDCG@10 \\
Beat detection & 0 & 4 & 4 & SciPy signal processing; event F1 \\
Link prediction & 0 & 4 & 4 & NetworkX features and logistic regression; ROC AUC \\
\midrule
Total & 22 & 14 & 30 & 36 conditions across eight families \\
\bottomrule
\end{tabularx}

\label{tab:catalog}
\end{table}

The catalog combines predictive learning, partition discovery, temporal
prediction, signal and image processing, and ranked retrieval. Its options
change representations, estimators, preprocessing, or computational operators.
Every outcome is produced by the documented native-library workflow on source
data. The source inventory, preparation, and all 48 workflow options appear in
Appendices~\ref{app:inventory}--\ref{app:options}. Each task records its source,
implementation, random seed, and evaluation cohort. Construction plans specify
the additional sources and configurations before their execution.

\begin{figure}[!ht]
\centering
\includegraphics[width=\linewidth]{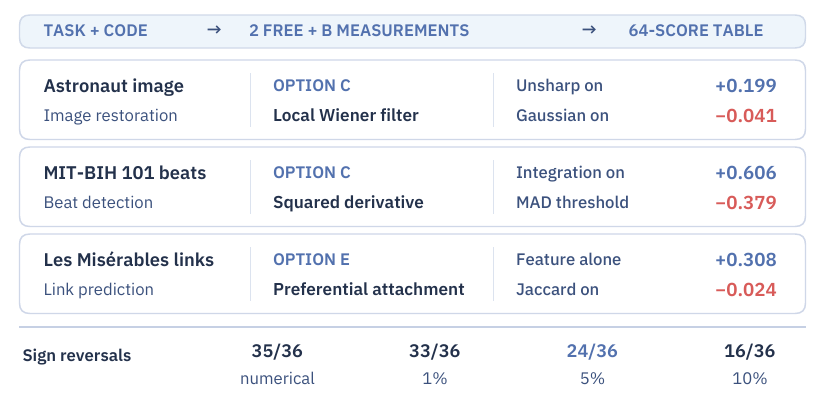}
\caption{Native conditional effects in three six-option tasks. Each row changes
one documented code option while fixing all other options in two backgrounds.
Values are evaluator-known changes in normalized task utility. The top strip shows
two free anchors, a budget of $B$ purchased outcomes, and a 64-score final
table. The bottom strip counts sign reversals over all 36 task conditions at
the indicated effect floors.}
\label{fig:catalog}
\end{figure}

\paragraph{Conditional structure across the catalog.}
Figure~\ref{fig:catalog} shows how one code option changes task utility
across two backgrounds in image restoration, beat detection, and link
prediction. Across the catalog, 24 of 36 conditions contain a sign reversal
with both directions above 5\% of the response range. The $10^{-12}$ numerical
floor yields 35 conditions; 1\% and 10\% floors yield 33 and 16. Flat exports
provide all 3,392 conditional effects and 342 mean pair interactions for reuse.

\paragraph{Data access and native evaluation.}
Training partitions fit supervised transforms and models, while feature-identical
rows remain grouped across splits. Clustering labels enter the external metric.
Forecasts use preceding observations with parameters fitted on the training
prefix. Restoration retains clean references at the evaluator, and retrieval
uses original relevance annotations with query-copy exclusion for ArguAna.
Beat detection uses archived MIT-BIH waveforms and event
annotations~\citep{mitbih,physionet}. Link prediction computes features on an
observed graph, fits calibration labels, and evaluates disjoint test pairs.
Appendix~\ref{app:preparation} specifies access rules and scoring cohorts.

\paragraph{Complete references and reproducible episodes.}
Cold execution checks all configuration predictions against their recorded
hashes. Separately implemented native metrics and target-perturbation checks
verify the scoring and evaluation boundaries. The prepared-input pack replays
all \CatalogConfigurations{} records from the prepared-input archive in
\CatalogReplaySeconds{} seconds with one computation thread. The standalone
episode engine agrees with \CatalogScoreChecks{} stored numerical-control
records. These checks establish a common reference for repeated evaluation.

An \emph{information budget} caps purchased outcomes. Agents view public tasks
and cached feedback; the evaluator retains complete references for matched
policy comparisons.

\section{Experimental setup}
\paragraph{Shared-estimator cross-comparison.}
For task $t$, acquisition policy $\pi$ produces an observation set
$O_{t,\pi}$. Let $\mathcal R_t(\pi,g)$ denote effect recovery when estimator
$g$ fits those observations. With $g$ fixed, contrasts across policies compare
acquired observation sets. With $O_{t,\pi}$ fixed, contrasts across estimators
compare reconstruction rules. 
\paragraph{Numerical reference methods.}
Main-effect ridge (D1) models additive component contributions; pair-effect
ridge (D2) additionally models pairwise interactions. Sequential ridge designs
spread measurements over these component features. Random design uses 20 seeds
with the pair estimator. A Gaussian-process baseline estimates a constant mean
and models similarity through the number of differing options~\citep{gpml}.
Regularization and kernel parameters minimize leave-one-out prediction error
on observed values. Grid GP and Effect GP select measurements to reduce
uncertainty in scores and conditional effects, respectively~\citep{cohnactive}.
All predictors retain measured values. Appendix~\ref{app:methods} gives the
feature maps, covariance, and acquisition criteria.

We additionally evaluate public-code equivalences for the original six-factor
variants. For example, a filter-width option is inactive when its filter is off.
An \emph{equivalence class} contains configurations with identical native
behavior. The control measures one representative per class and assigns its
prediction to every class member.
The trajectory records purchased measurements and inferred aliases separately.
Every equivalence is checked against native prediction hashes and scores.
Ridge controls, GP controls, and code-equivalence controls follow recorded
development stages. The protocol history and exact method grids appear in
Appendix~\ref{app:protocol}.

\paragraph{Agents and budgets.}
Numerical controls cover all eight workflow families. The 108-episode core agent
study uses Spambase, Concrete, Wine, CO$_2$, Astronaut, and SciFact, the six
one-per-family sources frozen with the initial four-factor agent study.
Matched controls use these same sources. Four-factor evaluation uses the API identifiers
\texttt{deepseek-v4-flash} and \texttt{deepseek-v4-pro} at $B=4,8$ for
72 episodes across the original and additional cohorts. We abbreviate these
identifiers as Flash and Pro. Both identifiers belong to one DeepSeek model
family. The evaluated tasks use binary switches in CPU-executable workflows.
Both models receive workflow code and input summaries,
reasoning enabled at high effort, a 32,768-token output cap, and a disclosed
1,800-second session deadline. The four tools expose current evidence, purchase
masks, execute numerical Python, and register a final prediction artifact. A maximum of
20 calculation calls applies, with 30 CPU seconds and 45 wall seconds per call.
NumPy, SciPy, scikit-learn, and public main-effect and pair-effect ridge helpers
are available.

The information study contains 36 Flash episodes at $B=20$ on six paired
six-factor workflows. Full-code and opaque views share tools, helpers, and
utility receipts; the full view adds code and factor semantics. Four-factor
evaluation contains three executions per source, model identifier, and budget
across dated cohorts. The $B=32$ beat-detection and graph extension contains
eight Flash attempts and six delivered tables. Exact
prompts, source pairs, and chronology appear in
Appendices~\ref{app:interface}--\ref{app:protocol}.

\pagebreak
\paragraph{Artifact-interface diagnostics.}
An \emph{artifact interface} receives a complete mask-to-utility table or a
request to fit a named estimator. Four six-factor sources with verified code equivalences
receive matched Flash episodes at $B=20$ under a typed table argument. Both
arms accept complete objects and report rules before measurement; the menu
adds GP and agent-supplied estimator rules to ridge fitting. A fresh inference
diagnostic holds each ridge-arm episode's 22 observations fixed and requires
a named estimator with submitted rules. Exact contracts and results appear in
Appendix~\ref{app:typed_interface}.

\paragraph{Prospective source cohort.}
A typed-interface study pairs Flash and Pro at $B=8$ on 12 four-factor
sources. The frozen plan selects the alphabetically first agent-unevaluated source ID
in each original family and retains the six core sources. Both models receive identical public tasks and
can submit tables or invoke ridge and GP helpers. Source-paired comparisons
fit a common GP to agent-purchased observations and frozen D2 observations.
The GP uses purchased masks and utilities; agent tables may also encode
public workflow code and sandbox calculations. The $R_{\rm all}-R_{\rm GP}$
contrast measures final-prediction gain over a fixed numerical baseline at
the same purchased masks. Delivery, all-attempt recovery, and delivered-table
recovery remain separate
(Appendix~\ref{app:typed_sources}).

\paragraph{Aggregation and delivery.}
Numerical controls and core cohorts average repetitions within source,
sources within family, and then weight families equally. We call this
\emph{family-macro averaging}. The prospective cohort averages sources equally;
$R_{\rm del}$ averages delivered sources; bootstrap ranges resample sources.
Masks and random seeds remain repeated observations within a source.
Four-factor evaluation delivers 63 of 72 artifacts. Every attempt retains its
delivery status, and missing delivery receives zero all-attempt recovery.
Appendix~\ref{app:protocol} gives transport accounting, and
Appendix~\ref{app:agentresults} reports all core episode outcomes.

\section{Results}
\subsection{Optimization success and intervention knowledge are distinct targets}
Exact configuration choice can coexist with arbitrarily small effect recovery,
even when both free anchors match. Appendix~\ref{app:metrics} gives a compact
construction. Native measurements show how the two targets separate in practice.

\begin{figure}[!ht]
\centering
\includegraphics[width=\linewidth]{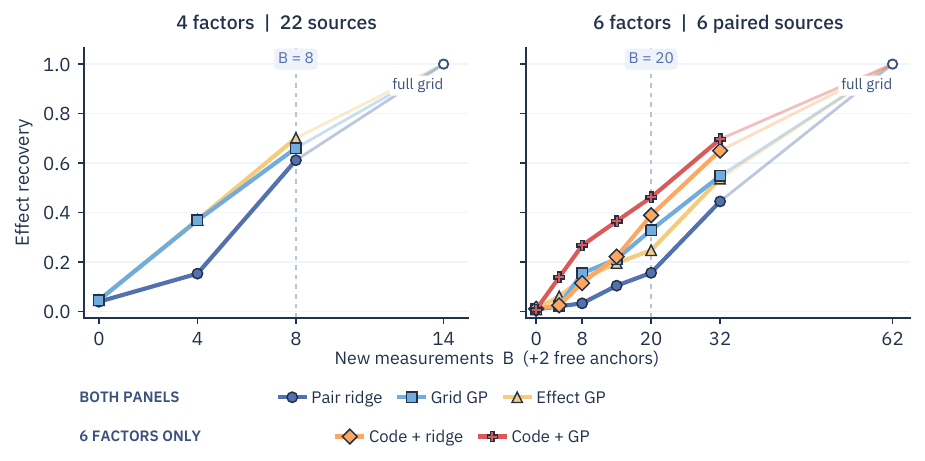}
\caption{Family-macro conditional-effect recovery across measurement budgets.
The left panel contains 22 four-factor sources; the right panel contains six
paired six-factor sources. Vertical guides mark the $B=8$ and $B=20$
comparison budgets. Pale final segments end at complete configuration
enumeration, with $B=14$ and $B=62$. Code-equivalence methods appear only
on the six-factor track and encode visible inactive-parameter rules.}
\label{fig:controls}
\end{figure}

Figure~\ref{fig:controls} compares recovery across budgets. On the 22 four-factor sources, pair-effect design/ridge at $B=8$ achieves 0.612
family-macro recovery, 15 exact configuration choices, and three strict
reconstructions. Thirteen instances combine exact choice with an effect error
above the strict tolerance. The effect-variance GP reaches 0.701 recovery,
16 exact choices, and one strict reconstruction at $\tau=0.1$. The 13 pair-ridge
instances that combine exact choice with a strict-effect error, together with
the construction in Appendix~\ref{app:metrics}, establish configuration choice
and conditional-effect accuracy as complementary evaluation targets.

Appendix~\ref{app:sensitivity} reports the complete strict-error tolerance
profile alongside native-unit results.

\subsection{Acquisition and estimator contrasts on common observations}
At $B=8$, Flash delivers recovery 0.632 and 0.621 across the original and
additional cohorts. Pro delivers 4/6 and 9/12 artifacts, with delivered-table
recovery 0.549 and 0.671. Among 18 Flash episodes, seven select an exact
optimum and one passes strict reconstruction. Pair ridge scores 0.627 on the
same six sources. Appendix Table~\ref{tab:models} separates every cohort's
delivery, all-attempt recovery, delivered recovery, and shared-GP fit.

At $B=8$, 32.6\% of original Flash effect edges have both endpoints measured;
67.4\% require prediction. Filling unmeasured utilities with the observation
mean gives normalized error 1.054 on these edges and recovery 0.287, versus
0.542 and 0.632 for delivered tables. The additional Flash cohort recovers
0.621 versus 0.213 for mean completion (Appendix~\ref{app:calibration}).

All 94 valid core final artifacts select public ridge helpers, with 77
pair-effect and 17 main-effect selections. Both helpers use the same
leave-one-out regularization grid as the shared D1 and D2 fits. Each table
equals its selected shared fit. At $B=8$, Flash selects D1/D2 in 2/10 original
and 2/4 additional episodes; the selected fit has higher same-observation
recovery in 7/12 and 5/6 episodes, respectively.
Table~\ref{tab:crossfit} crosses acquisition policies with common D1, D2, and
GP estimators at $B=8$ on the same six sources. Point estimates reorder
policies across estimators. Original Flash observations yield 0.698 recovery with the
shared GP versus 0.669 for D2 observations; under shared D2, the values are
0.550 and 0.627. The additional Flash cohort repeats this pattern with 0.720
versus 0.669 under GP and 0.537 versus 0.627 under D2. The pooled
Flash-minus-D2 acquisition contrast under this GP is
$+0.040$ (descriptive interval $[-0.039,0.132]$; four of six sources positive). The
source-level paired contrasts appear in Appendix~\ref{app:source_sensitivity}.

\begin{table}[!ht]
\caption{Core-cohort acquisition and reconstruction at $B=8$ on six matched four-factor
sources. Rows fix an acquisition policy; shared columns fit each estimator to
that policy's observations. Random designs average 20 seeds within source.
$R_{\rm all}$ assigns zero to missing artifacts, and shared fits use all
acquired evidence.}
\centering
{%
\fontsize{9}{11}\selectfont%
\setlength{\tabcolsep}{3pt}%
\begin{tabularx}{\linewidth}{@{}lRRRRR@{}}
\toprule
\rowcolor{WWBHeader}
\textbf{Acquisition} & \textbf{Records} & \textbf{$R_{\rm all}$} & \textbf{Shared D1} & \textbf{Shared D2} & \textbf{Shared GP} \\
\midrule
Flash, original & 12 & 0.632 & 0.604 & 0.550 & 0.698 \\
Pro, original & 6 & 0.366 & 0.628 & 0.567 & 0.692 \\
Flash, additional & 6 & 0.621 & 0.598 & 0.537 & 0.720 \\
Pro, additional & 12 & 0.503 & 0.564 & 0.670 & 0.667 \\
\midrule
D1 & 6 & 0.546 & 0.546 & 0.592 & 0.592 \\
D2 & 6 & 0.627 & 0.571 & 0.627 & 0.669 \\
Random D2 & 120 & 0.523 & 0.542 & 0.523 & 0.619 \\
GP-G & 6 & 0.666 & 0.564 & 0.633 & 0.666 \\
GP-E & 6 & 0.640 & 0.553 & 0.568 & 0.640 \\
\bottomrule
\end{tabularx}%
}

\label{tab:crossfit}
\end{table}

\paragraph{Prospective inference across 12 sources.}
Table~\ref{tab:crossfit} records selected core ridge fits. The typed cohort
in Table~\ref{tab:typed_sources_main} has a distinct prompt, execution date,
and direct-table contract. On the six recurring sources, the shared-GP
Flash-minus-D2 acquisition point estimates are $+0.040$ in core and $-0.051$
in typed runs.
Flash submits 12 direct tables and exceeds the same-observation GP on 11/12
sources, with mean $R_{\rm all}-R_{\rm GP}=+0.147$. Pro delivers 11/12 artifacts.
Its all-attempt contrast is $-0.001$; among delivered artifacts, its mean
contrast with the GP on those same 11 sources is $+0.063$.
\begin{table}[!ht]
\caption{Typed cohort on six new and six original sources at $B=8$.
$R_{\rm all}$ scores absent submissions as zero; $R_{\rm del}$ averages
delivered tables. $R_{\rm GP}$ uses the same observations. Gain is
$R_{\rm all}-R_{\rm GP}$. Ranges give descriptive 2.5th--97.5th
source-bootstrap percentiles.}
\centering
{\fontsize{8}{10}\selectfont\setlength{\tabcolsep}{1.7pt}%
\begin{tabularx}{\linewidth}{@{}llRRRRRR@{}}
\toprule
\rowcolor{WWBHeader}
\textbf{Sources} & \textbf{Model} & \textbf{Valid} & \textbf{$R_{\rm all}$} & \textbf{$R_{\rm del}$} & \textbf{$R_{\rm GP}$} & \textbf{Gain} & \textbf{Range} \\
\midrule
New & Flash & 6/6 & 0.869 & 0.869 & 0.719 & $+0.149$ & $[0.067,0.241]$ \\
New & Pro & 6/6 & 0.792 & 0.792 & 0.718 & $+0.074$ & $[0.006,0.160]$ \\
Original & Flash & 6/6 & 0.764 & 0.764 & 0.618 & $+0.146$ & $[-0.005,0.360]$ \\
Original & Pro & 5/6 & 0.590 & 0.708 & 0.666 & $-0.076$ & $[-0.392,0.215]$ \\
\midrule
All & Flash & 12/12 & 0.816 & 0.816 & 0.669 & $+0.147$ & $[0.055,0.261]$ \\
All & Pro & 11/12 & 0.691 & 0.754 & 0.692 & $-0.001$ & $[-0.177,0.153]$ \\
\bottomrule
\end{tabularx}}
\par\smallskip
{\footnotesize\raggedright A post-hoc oracle using the full reference selects
the best same-observation D1, D2, or GP fit per source. Flash retains gains
of 0.114 on new sources (6/6 positive) and 0.095 overall (10/12 positive).\par}
\label{tab:typed_sources_main}
\end{table}
The prespecified six-source new subgroup yields positive table-minus-GP
contrasts on 6/6 Flash sources and 4/6 Pro sources, with one Pro tie and one
negative contrast. Pro selects the named GP on Coffee, giving an exact tie.
With shared GP, agent-minus-D2 acquisition point estimates for (Flash, Pro)
are $(-0.024,-0.026)$ on new sources and $(-0.037,-0.014)$ across all 12.
The reported six- and twelve-source bootstrap ranges span zero
(Appendix~\ref{app:typed_sources}).

\subsection{Program rules and final predictions}
Across four code-visible sources, the typed arms deliver 8/8 direct tables;
seven satisfy every verified equivalence. The four paired menu-minus-ridge recovery
contrasts average $-0.092$, with one positive and three negative sources.
Table~\ref{tab:typed_interface} tracks rule reporting and final artifacts.

\begin{table}[!ht]
\caption{Rule reporting and final artifacts on four six-factor sources.
Rules passed counts
registered relations supplied to a named estimator in the final submission
call; direct-table submissions pass none. Clean tables satisfy all verified
equivalences.
}\centering
{\fontsize{8}{10}\selectfont\setlength{\tabcolsep}{2.5pt}%
\begin{tabularx}{\linewidth}{@{}lRRRRRR@{}}
\toprule
\rowcolor{WWBHeader}
\textbf{Study} & \textbf{Delivered} & \textbf{First rules} & \textbf{Final mode} & \textbf{\shortstack{Rules passed\\to named\\estimator}} & \textbf{Clean} & \textbf{Recovery} \\
\midrule
Typed ridge & 4/4 & 6/6 & 4 direct & 0 & 3/4 & 0.710 \\
Typed menu & 4/4 & 6/6 & 4 direct & 0 & 4/4 & 0.617 \\
\midrule
Fixed observations & 4/4 & 6/6 & 3 GP, 1 D1 & 6/6 & 4/4 & 0.682 \\
\bottomrule
\end{tabularx}}
\label{tab:typed_interface}
\end{table}

Four fresh agents pass all six
rules to named estimators; all 96 equivalent pairs agree. Their mean recovery
is 0.682 versus 0.710 for the separate typed direct-table arms. On fixed
ridge-arm observations, verified rules raise D2 recovery from 0.319 to 0.601
and GP recovery from 0.509 to 0.666 (Appendix~\ref{app:typed_interface}).
The six-workflow controls jointly apply class-aware acquisition and prediction
(Appendices~\ref{app:program_equiv_controls} and~\ref{app:native_sensitivity}).
In a separate eight-episode rule-reporting diagnostic, agents identify all
registered code relations. Their ridge-helper tables show native-equivalence
violations. A class-consistent control scores 0.507 versus 0.338 for the
delivered tables (Appendix~\ref{app:structure_probe}).

\section{Conclusion}
WhatWorkedBench grades budgeted response surfaces against exhaustive CPU
references for conditional effects, pair interactions, configuration choice,
and delivery. Its \CatalogConditions{} conditions span \CatalogSources{}
sources and \CatalogFamilies{} workflow families, with \CatalogConfigurations{}
indexed configuration records and 4,206 numerical controls. At eight purchases,
pair-effect ridge selects the exact optimum yet exceeds the $\tau=0.1$ strict
effect-error tolerance on 13 of 22 four-factor sources. Shared-estimator
comparisons report acquisition and reconstruction on common observations.
In the 12-source typed study of DeepSeek Flash and Pro, 23/24 episodes deliver
final artifacts. Flash gains 0.147 recovery over the same-observation GP across
all 12 sources; Pro has a $-0.001$ all-attempt contrast and a $+0.063$
delivered-only contrast. Six new sources yield Flash and Pro gains of 0.149
and 0.074. Seven of eight typed six-factor tables satisfy verified
equivalences. Four fresh agents deliver rule-consistent tables.

\newpage
\section*{Reproducibility statement}
The public benchmark repository is available at
\url{https://github.com/EthanNing/WhatWorkedBench}. It includes public tasks, evaluator-only
references, versioned episode semantics, exact native operations, prepared
inputs, numerical controls, and machine-readable audits. The task pack's
standard-library engine reproduces \CatalogScoreChecks{} stored score records
in a standalone process. The prepared-input pack reproduces all
\CatalogConfigurations{} native records from the public repository. Source
acquisition and preparation retain separate manifests and code identities.

\section*{AI use statement}
An AI coding assistant contributed to topic and literature exploration,
workflow-option design, experiment protocols, data preparation, implementation,
numerical analysis, figures, and drafting. DeepSeek models were additionally
used as evaluated experimental agents. Verification used cold native replay,
separately implemented numerical metrics, and recorded source and code
identities. The study preserves the generated implementations, analysis
scripts, experiment plans, and numerical audit results.

\section*{Ethics statement}
This work uses publicly available research archives with their accompanying
source terms and citations. The ECG tasks evaluate computational event detection
on archived waveforms. Prepared inputs contain the signal data and reference
events used for scoring. Evaluation references are held by the trusted host,
and agents receive task descriptions and budgeted aggregate observations.

\appendix
\raggedbottom
\section{Benchmark design and related evaluations}
\label{app:related}
This appendix retains source and option definitions, metric properties,
numerical methods, agent prompts, and comparisons needed to interpret the
results. Complete configuration, prediction, and trajectory records remain in
the accompanying anonymous supplement. Displayed statistics are rounded after computation.

\paragraph{Interactive experimental discovery.}
ScienceWorld tests grounded scientific reasoning in an interactive text
environment~\citep{scienceworld}. BoxingGym directly evaluates experimental design and acquired predictive
knowledge through expected information gain, prediction errors, and the
predictions enabled by an agent's explanation~\citep{boxinggym}.
DiscoveryWorld evaluates complete scientific investigations in virtual worlds,
including hypothesis formation, experiments, and conclusions~\citep{discoveryworld}.
Gravity-Bench budgets observations of gravitational systems and grades numerical
answers about concealed physics~\citep{gravitybench}. Science-Gym couples data
collection with equation discovery across physical and epidemiological
simulations~\citep{sciencegym}. SciGym evaluates reconstruction of biological
systems~\citep{scigym}, while DiscoverPhysics scores executable inferred laws
and explanations in noncanonical physical worlds~\citep{discoverphysics}.
CausalGame evaluates active experimental design and causal explanation in games
with selection bias, measurement error, and hidden confounding~\citep{causalgame}.
These works establish interactive discovery and prediction as existing
benchmark targets. WhatWorkedBench's reference consists of exhaustive native
outcomes for documented component interventions, supporting error measurement
on every conditional component effect and reuse of identical observations
across reconstruction methods.

\paragraph{Understanding and attribution.}
Science sandboxes explicitly distinguish successful optimization from learning
system rules, using regulatory-genomics and protein-fitness settings~\citep{sciencesandboxes}.
CAFE registers pipeline components as factors, executes factorial designs, and
uses mixed-effects models to attribute rated answer quality to components and
interactions~\citep{cafe}. The petri-bench technical report evaluates identification
of hidden parameter changes and, at its interaction tier, the sign of a
controlled two-factor contrast; scoring also checks the supporting experimental
procedure~\citep{petribench}. WhatWorkedBench grades the entire conditional-effect
field implied by a submitted response surface against native metric outcomes.
Its cross-comparison contrasts measurement policies under shared estimators and
estimators on the same acquired set. Factorial attribution and the distinction
between performance and understanding supply foundations for this particular
measurement design.
DCP grades research-result recovery and calibrated feedback effects under
matched controls~\citep{dcp}. Our delivered response surface makes
intervention knowledge independently gradable.

\paragraph{Research workflows and scientific tools.}
DeepResearchGym provides a reproducible search environment and evaluates
research reports for information-need alignment, retrieval faithfulness, and
quality~\citep{deepresearchgym}; our fixed experimental environment evaluates
quantitative predictions of component interventions.
MLAgentBench and MLE-bench evaluate executable machine-learning solutions and
model performance~\citep{mlagentbench,mlebench}. BLADE scores analysis decisions against expert analyses~\citep{blade};
DiscoveryBench evaluates hypotheses about relationships in supplied datasets
through structured facets~\citep{discoverybench}. ScienceAgentBench evaluates
executable programs for scientific tasks drawn from publications~\citep{scienceagentbench}.
EXP-Bench covers design, implementation, execution, and conclusions for
paper-derived experiments~\citep{expbench}. ResearchGym preserves datasets,
evaluators, and baselines while withholding proposed methods~\citep{researchgym};
InnovatorBench evaluates runnable contributions to LLM research~\citep{innovatorbench}.
Closed-loop molecular Auto Research evaluates feature and model interventions
under held-out scoring~\citep{molecularresearch}.
AstaBench evaluates literature, analysis, coding, and research tasks with
reproducible tools and agent baselines~\citep{astabench}.
RExBench tests expert-specified extensions to existing research code~\citep{rexbench}.
FIRE-Bench evaluates the rediscovery of established findings~\citep{firebench}.
AgentActionBench evaluates reproduction traces using paper-specific rubrics~\citep{agentactionbench},
and ScienceBoard evaluates scientific workflows using professional software~\citep{scienceboard}.
SciAgentGym evaluates scientific tool orchestration, and SciAgentArena provides
interactive research tasks with stepwise verification~\citep{sciagentgym,sciagentarena}.
These targets concern successful research actions and outputs. The finite
workflow intervention reference here permits exhaustive assessment of what an
agent can predict after purchasing only part of the available evidence.

\paragraph{Planning and learned artifacts.}
AbGen, AblationBench, and SCOPE evaluate proposed ablations or experiment
plans using paper-derived references and expert or model judgments~\citep{abgen,ablationbench,scope}.
WhatWorkedBench adds execution-backed measurements and quantitative predictions
to the evaluated episode. HExA learns composable skills through active
experimentation in Interphyre~\citep{hexa}; SkillLearnBench evaluates procedural
skills at the levels of skill quality, execution trajectory, and task outcome~\citep{skilllearnbench}.
Here the delivered knowledge artifact is a complete numerical response surface.
The Implementation Lottery measures outcome variation across implementations
of one research idea~\citep{implementationlottery}. Fixed workflow implementations
give this benchmark stable references for each documented code-option change.

\paragraph{Design and response-surface foundations.}
Computer experiments connect budgeted code execution to statistical prediction
of computational responses~\citep{sacks1989computer}.
Bayesian experimental design chooses observations to inform a declared target~\citep{modernbed}.
Deep Adaptive Design learns policies for sequential information acquisition~\citep{dad}.
Predictive-variance active learning supplies the basis for our grid and linear
contrast acquisition criteria~\citep{cohnactive}; the Gaussian-process reference
uses standard conditioning and covariance calculations~\citep{gpml}.
Functional ANOVA estimates hyperparameter importance and interaction contributions
from surrogate models~\citep{fanova}. HPOBench provides reproducible optimization
objectives, while LLAMBO uses language models for initialization, surrogate
modeling, and candidate selection~\citep{hpobench,llambo}.
Our exhaustive tables provide observed references for conditional-effect error,
allowing different acquisition and inference methods to share one evaluation
object. The complete binary-cube representation follows standard Fourier
analysis on the Boolean cube~\citep{booleananalysis}.

\section{Source inventory}
\label{app:inventory}
The catalog contains 30 distinct sources and 36 task conditions. The table
links every source identifier to its recorded source location and reports the
prepared input and evaluation size. The six entries marked $d=4/6$ have paired
dimensional variants. Their E=F=0 subcubes reproduce the original four-factor
outcomes, accounting for 96 repeated configuration records across the two tracks.

\begin{center}
{%
\fontsize{8}{10}\selectfont%
\setlength{\tabcolsep}{3pt}%
\begin{tabularx}{\linewidth}{@{}llllR@{}}
\toprule
\rowcolor{WWBHeader}
\textbf{Family} & \textbf{Source ID} & \textbf{Prepared input} & \textbf{Evaluation} & \textbf{$d$} \\
\midrule
Classification & \href{https://archive.ics.uci.edu/dataset/267/banknote+authentication}{banknote} & 943 rows $\times$ 4 & 405 rows & 4 \\
 & \href{https://archive.ics.uci.edu/dataset/17/breast+cancer+wisconsin+diagnostic}{breast\_diagnostic} & 398 rows $\times$ 30 & 171 rows & 4 \\
 & \href{https://archive.ics.uci.edu/dataset/52/ionosphere}{ionosphere} & 245 rows $\times$ 34 & 105 rows & 4 \\
 & \href{https://archive.ics.uci.edu/dataset/94/spambase}{spambase} & 2,947 rows $\times$ 57 & 1,263 rows & 4/6 \\
\midrule
Regression & \href{https://archive.ics.uci.edu/dataset/291/airfoil+self+noise}{airfoil} & 1,052 rows $\times$ 5 & 451 rows & 4 \\
 & \href{https://archive.ics.uci.edu/dataset/165/concrete+compressive+strength}{concrete} & 702 rows $\times$ 8 & 303 rows & 4/6 \\
 & \href{https://archive.ics.uci.edu/dataset/242/energy+efficiency}{energy\_efficiency} & 537 rows $\times$ 8 & 231 rows & 4 \\
 & \href{https://archive.ics.uci.edu/dataset/243/yacht+hydrodynamics}{yacht} & 215 rows $\times$ 6 & 93 rows & 4 \\
\midrule
Clustering & \href{https://archive.ics.uci.edu/dataset/602/dry+bean+dataset}{dry\_bean} & 13,543 rows $\times$ 16 & 7 reference classes & 4 \\
 & \href{https://archive.ics.uci.edu/dataset/42/glass+identification}{glass} & 213 rows $\times$ 9 & 6 reference classes & 4 \\
 & \href{https://archive.ics.uci.edu/dataset/236/seeds}{seeds} & 210 rows $\times$ 7 & 3 reference classes & 4 \\
 & \href{https://archive.ics.uci.edu/dataset/109/wine}{wine} & 178 rows $\times$ 13 & 3 reference classes & 4/6 \\
\midrule
Forecasting & \href{https://raw.githubusercontent.com/statsmodels/statsmodels/v0.14.6/statsmodels/datasets/co2/co2.csv}{co2} & 1,713 training points & 571 next points & 4/6 \\
 & \href{https://raw.githubusercontent.com/statsmodels/statsmodels/v0.14.6/statsmodels/datasets/elnino/elnino.csv}{elnino} & 549 training points & 183 next points & 4 \\
 & \href{https://raw.githubusercontent.com/statsmodels/statsmodels/v0.14.6/statsmodels/datasets/nile/nile.csv}{nile} & 75 training points & 25 next points & 4 \\
 & \href{https://raw.githubusercontent.com/statsmodels/statsmodels/v0.14.6/statsmodels/datasets/sunspots/sunspots.csv}{sunspots} & 231 training points & 78 next points & 4 \\
\midrule
Image restoration & \href{https://raw.githubusercontent.com/scikit-image/scikit-image/v0.26.0/src/skimage/data/astronaut.png}{astronaut} & 256 $\times$ 256 pixels & Clean reference & 4/6 \\
 & \href{https://raw.githubusercontent.com/scikit-image/scikit-image/v0.26.0/src/skimage/data/coffee.png}{coffee} & 171 $\times$ 256 pixels & Clean reference & 4 \\
 & \href{https://raw.githubusercontent.com/scikit-image/scikit-image/v0.26.0/src/skimage/data/hubble_deep_field.jpg}{hubble\_deep\_field} & 223 $\times$ 256 pixels & Clean reference & 4 \\
 & \href{https://raw.githubusercontent.com/scikit-image/scikit-image/v0.26.0/src/skimage/data/rocket.jpg}{rocket} & 171 $\times$ 256 pixels & Clean reference & 4 \\
\midrule
Retrieval & \href{https://www.arguana.com/data}{arguana} & 8,567 documents & 64 queries & 4 \\
 & \href{https://github.com/allenai/scifact}{scifact} & 5,183 documents & 64 queries & 4/6 \\
\midrule
Beat detection & \href{https://physionet.org/content/mitdb/1.0.0/}{mitdb\_100} & 108,000 at 360 Hz & 369 interior events & 6 \\
 & \href{https://physionet.org/content/mitdb/1.0.0/}{mitdb\_101} & 108,000 at 360 Hz & 340 interior events & 6 \\
 & \href{https://physionet.org/content/mitdb/1.0.0/}{mitdb\_102} & 108,000 at 360 Hz & 364 interior events & 6 \\
 & \href{https://physionet.org/content/mitdb/1.0.0/}{mitdb\_103} & 108,000 at 360 Hz & 352 interior events & 6 \\
\midrule
Link prediction & \href{https://websites.umich.edu/~mejn/netdata/}{graph\_adjnoun} & 112 nodes, 255 edges & 170 cal., 170 test & 6 \\
 & \href{https://websites.umich.edu/~mejn/netdata/}{graph\_dolphins} & 62 nodes, 95 edges & 64 cal., 64 test & 6 \\
 & \href{https://websites.umich.edu/~mejn/netdata/}{graph\_football} & 115 nodes, 367 edges & 246 cal., 246 test & 6 \\
 & \href{https://websites.umich.edu/~mejn/netdata/}{graph\_lesmis} & 77 nodes, 152 edges & 102 cal., 102 test & 6 \\
\bottomrule
\end{tabularx}%
}

\end{center}

\paragraph{Source terms and attribution.}
The UCI tabular sources retain CC-BY-4.0 attribution in their manifests.
The statsmodels 0.14.6 time-series metadata identify their data as public domain.
The scikit-image collection retains source-specific attribution for the NASA
Astronaut image, Rachel Michetti's CC0 Coffee image, Hubble Deep Field, and the
SpaceX Rocket image. SciFact claims and annotations use CC-BY-4.0, with corpus
terms recorded as ODC-By-1.0. ArguAna preserves the original authors'
\href{https://zenodo.org/records/3973258}{CC-BY-4.0 deposit}.
MIT-BIH uses Open Data Commons Attribution v1.0~\citep{mitbih,physionet}.

The graph archive attributes Les Miserables coappearance to D. E. Knuth,
\emph{The Stanford GraphBase} (1993); David Copperfield word adjacency to
M. E. J. Newman, \emph{Physical Review E} 74, 036104 (2006); college football
to M. Girvan and M. E. J. Newman, \emph{PNAS} 99, 7821--7826 (2002); and dolphin
associations to D. Lusseau et al., \emph{Behavioral Ecology and Sociobiology}
54, 396--405 (2003). Its source page permits scientific use. Complete source
terms, original row counts, duplicate removals, and acquisition metadata remain
with each source manifest.

Photographs share the scikit-image collection, time series share the statsmodels
distribution, ECG records share MIT-BIH, and graph instances share an archive.
Source-level reporting retains these relationships and merges paired variants
when counting distinct sources.

\section{Data preparation and evaluation boundaries}
\label{app:preparation}
\subsection{Tabular classification, regression, and clustering}
The UCI metadata define feature roles, index columns, and target columns.
Acquisition checks the dataset identifier and feature count against these
metadata. Regression uses the designated numerical target. Energy Efficiency
uses heating load Y1, with the secondary target excluded from the features.
Classification and clustering map sorted original class labels to integer
identifiers. All retained feature and target values are finite.

Preparation removes exact duplicate feature--label rows while preserving the
order of first occurrence. Supervised datasets then group rows by identical
feature vectors. A single \texttt{GroupShuffleSplit} with evaluation fraction
0.3 and seed 20260907 assigns complete groups to training or evaluation. Both
parts of every classification split contain the same class set. The split is
fixed across masks. Feature transforms, feature selection, PCA, and predictive
models fit on training data and transform the evaluation features afterward.

Clustering operates on all deduplicated feature rows. The known number of
classes fixes $k$. K-means fits the feature representation for each mask, and
the original class labels enter the external NMI calculation. The evaluation
therefore measures agreement between the resulting partition and the archived
class partition on the fixed data.

\subsection{Chronological forecasting}
The time-series files come from statsmodels version 0.14.6. El Ni\~no year rows
are expanded in calendar-month order. Other series preserve their source time
order. A missing value receives the latest preceding finite value. CO2 has
59 such replacements, and the other three series have zero. The training cut
is $\lfloor0.75N\rfloor$, and the suffix supplies the evaluation points.

All configurations start their training examples at index 12 so the one-lag
and 12-lag designs use the same fitting times. The predictor fits once on the
training prefix. Evaluation at time $t$ uses the realized observations preceding
$t$, giving a sequence of one-step predictions with fixed fitted parameters.
The six-factor first-difference option applies after the optional logarithm and
restores each prediction using the preceding realized level. This contract
defines the temporal information available to every mask.

\subsection{Image preparation and restoration}
Image bytes are checked against the scikit-image 0.26.0 registry. Pillow converts
the source to grayscale, uses a Lanczos thumbnail with maximum width and height
256, and scales intensities by 255. The resulting array supplies the clean
reference. A NumPy generator with seed 20260907 adds independent Gaussian noise
with standard deviation 0.1, followed by clipping to $[0,1]$. Each source uses
the same prepared noisy array across all masks. Native filters operate on that
array, and the clean array enters only the external pixel-RMSE calculation.

\subsection{Retrieval corpora and query cohorts}
The retrieval sources use BEIR corpus exports~\citep{beir}.
SciFact uses 5,183 corpus documents and the hash-first 64 original training
queries. ArguAna uses 8,567 corpus documents and the first 64 prepared G queries
from its original training partition. Original relevance annotations supply
the target. Corpus documents are sorted by document identifier before vectorizer
fitting. Query vectors use the corpus-fitted vocabulary and weighting.
Stable descending score order determines the top ten documents.

ArguAna removes every document whose normalized text matches the query text,
under every mask. The recorded split audit checks query text, counter-text,
reversed pairs, and debate-group relationships. The prepared query cohorts and
the source terms are preserved in the source manifests. Retrieval utility
averages the same 64 query-level NDCG values for every configuration.

\subsection{Archived signal records}
MIT-BIH records 100, 101, 102, and 103 contribute samples $[0,108000)$ from channel
zero at 360 Hz. The first channel's physical units are millivolts. WFDB reads
the waveform and original event annotations. The QRS label-code set is
$\{1,2,3,4,5,6,7,8,9,10,11,12,13,25,30,31,34,35,38,41\}$, following the source labels.

Native prediction receives the waveform and sample rate. Scoring restricts
predicted and reference events to the interior $[1,299)$-second interval.
The source inventory reports the interior event counts used for scoring. The filters
use offline forward--backward filtering. Consequently, this task evaluates
event localization within a fixed archived segment. Event matching uses a
100-ms tolerance, with 50-ms and 150-ms checks retained for sensitivity.

\subsection{Graph construction and pair partitions}
The source graphs are converted to simple, undirected, unweighted graphs.
Self-loops are removed and node identifiers are mapped to consecutive integers.
A source-specific fixed seed shuffles the original edges. Calibration and test
each receive $\operatorname{round}(0.2|E|)$ positive edges. The remaining edges
form the observed graph, giving approximately 60/20/20 proportions. The shuffled
sequence assigns observed edges first, then calibration positives, then test
positives. Partition sizes and source-specific seeds accompany the prepared arrays.

Calibration and test negatives are disjoint balanced samples from the original
graph's nonedges. Every feature for both sets is computed on the shared observed
graph. Calibration labels fit logistic regression. Test labels supply the
external ROC AUC. This creates a reproducible finite link-prediction problem
with separate observed, calibration-positive, and test-positive edges.

\section{Workflow options and fixed numerical settings}
\label{app:options}
Mask bits run left to right as A, B, C, D, E, F. The four-factor track uses A--D.
The paired track adds E and F with the original options and order preserved.
Every binary mask is legal. The tables specify both settings for each option, followed by operation order and numerical details.
The public task files contain the corresponding executable operation definitions.

\newcommand{\optionhead}{\rowcolor{WWBHeader}\textbf{Bit} & \textbf{Setting 0} & \textbf{Setting 1} \\\midrule}

\subsection{Classification}
\begin{center}\small
\begin{tabularx}{\linewidth}{@{}lp{5.4cm}L@{}}
\toprule\optionhead
A & Original features & $\operatorname{sign}(x)\log(1+|x|)$ \\
B & Original feature scale & Training-fitted standard scaling \\
C & Uniform class weights & Balanced class weights \\
D & Logistic regression $C=1$ & Logistic regression $C=0.05$ \\
E & All input features & Training-fitted ANOVA selection at the 50th percentile \\
F & Input representation & PCA retaining 95\% variance after optional selection \\
\bottomrule
\end{tabularx}
\end{center}
Logistic regression uses the LBFGS solver, 3,000 maximum iterations, and random
state 17. Standard scaling precedes selection and PCA. PCA uses the full SVD.
Class balancing follows the training class frequencies.

\subsection{Regression}
\begin{center}\small
\begin{tabularx}{\linewidth}{@{}lp{5.4cm}L@{}}
\toprule\optionhead
A & Original target & Training-target \texttt{log1p} with \texttt{expm1} inverse \\
B & Original feature scale & Training-fitted feature scaling \\
C & Original features & Degree-two polynomial features without a bias column \\
D & Ridge $\alpha=1$ & Ridge $\alpha=10$ \\
E & Full degree-two polynomial terms & Interaction-only polynomial terms when C is enabled \\
F & Standard scaling when B is enabled & Robust scaling when B is enabled \\
\bottomrule
\end{tabularx}
\end{center}
Scaling precedes polynomial construction. Ridge uses the SVD solver.
The target transform wraps the complete regressor. E is inactive at C=0,
and F is inactive at B=0.

\subsection{Clustering}
\begin{center}\small
\begin{tabularx}{\linewidth}{@{}lp{5.4cm}L@{}}
\toprule\optionhead
A & Original features & Signed logarithmic feature transform \\
B & Original feature scale & Feature-only standard scaling \\
C & All input features & PCA retaining 95\% variance \\
D & One k-means initialization & Ten k-means initializations \\
E & Unwhitened PCA coordinates & Whitened coordinates when C is enabled \\
F & K-means++ initialization & Random initialization \\
\bottomrule
\end{tabularx}
\end{center}
K-means uses Lloyd's algorithm, a maximum of 300 iterations, and random state 17.
The known class count determines $k$. PCA uses the full SVD. E is inactive at C=0.

\subsection{Forecasting}
\begin{center}\small
\begin{tabularx}{\linewidth}{@{}lp{5.4cm}L@{}}
\toprule\optionhead
A & Lag features & Lag features and a linear time trend $t/\mathrm{cut}$ \\
B & Lag 1 & Lags 1 through 12 \\
C & Original series levels & \texttt{log1p} levels with \texttt{expm1} inverse \\
D & Ridge $\alpha=1$ & Ridge $\alpha=10$ \\
E & Training-fitted standard scaling & Training-fitted robust scaling \\
F & Model series levels & Model first differences and restore the previous level \\
\bottomrule
\end{tabularx}
\end{center}
Ridge uses the SVD solver. Optional differencing follows the series transform.
The first difference at index zero is set to zero, and fitting starts at index
12 for every mask. Forecast inversion restores differenced levels before the
optional exponential inverse.

\subsection{Image restoration}
\begin{center}\small
\begin{tabularx}{\linewidth}{@{}lp{5.4cm}L@{}}
\toprule\optionhead
A & Pass the image onward & Median filter with a $3\times3$ window \\
B & Pass the image onward & Gaussian filter with $\sigma=0.8$ \\
C & Pass the image onward & Wiener filter with a $5\times5$ neighborhood \\
D & Pass the image onward & Unsharp mask with Gaussian $\sigma=1$, amount 0.5 \\
E & Gaussian $\sigma=0.8$ when B is enabled & Gaussian $\sigma=1.6$ when B is enabled \\
F & Median window $3\times3$ when A is enabled & Median window $5\times5$ when A is enabled \\
\bottomrule
\end{tabularx}
\end{center}
Filters execute in A, B, C, D order and the final image is clipped to $[0,1]$.
Median and Gaussian filters use reflected boundaries. Wiener noise power is
estimated by the library from the current image. E is inactive at B=0 and
F is inactive at A=0.

\subsection{Retrieval}
\begin{center}\small
\begin{tabularx}{\linewidth}{@{}lp{5.4cm}L@{}}
\toprule\optionhead
A & Unigrams & Unigrams and bigrams \\
B & Retain stopwords & Remove the scikit-learn English stopword set \\
C & Raw term frequency & Sublinear term frequency \\
D & Unit term weights & Inverse-document-frequency weighting \\
E & Frequency counts & Binary term counts \\
F & L2 vector normalization & Unnormalized document and query vectors \\
\bottomrule
\end{tabularx}
\end{center}
TF--IDF vectors use float64. The vectorizer fits on corpus text and transforms
queries. Dot products supply ranking scores. Binary counts make the sublinear
transform inactive. Document sorting and stable ranking fix score ties.

\subsection{Beat detection}
\begin{center}\small
\begin{tabularx}{\linewidth}{@{}lp{5.4cm}L@{}}
\toprule\optionhead
A & Median-centered waveform & Additional second-order 0.5-Hz high-pass filter \\
B & Current waveform & Additional second-order 35-Hz low-pass filter \\
C & Absolute-amplitude envelope & Squared first-difference envelope \\
D & Current envelope & 120-ms moving integration \\
E & Mean plus one standard deviation & Median plus three scaled MAD units \\
F & 200-ms minimum peak distance & 300-ms minimum peak distance \\
\bottomrule
\end{tabularx}
\end{center}
The MAD scale factor is 1.4826. Butterworth filters use second-order sections
and forward--backward filtering. Integration uses nearest-edge extension.
Each candidate aligns to the largest absolute filtered amplitude within 75 ms.
Duplicate samples merge before scoring.

\subsection{Link prediction}
\begin{center}\small
\begin{tabularx}{\linewidth}{@{}lp{5.4cm}L@{}}
\toprule\optionhead
A & Feature absent & Common-neighbor count \\
B & Feature absent & Jaccard coefficient \\
C & Feature absent & Adamic--Adar index \\
D & Feature absent & Resource-allocation index \\
E & Feature absent & Preferential-attachment score \\
F & Original feature scales & Calibration-fitted standard scaling \\
\bottomrule
\end{tabularx}
\end{center}
A mask with all five feature bits zero supplies a zero feature column and fits
an intercept-only model. Logistic regression uses $C=1$, LBFGS, 2,000 maximum
iterations, and random state 20260907. The positive-class probability supplies
the ranking score for ROC AUC.

\section{Response scales and conditional structure}
\label{app:conditions}
The table summarizes response widths $\max_z u_t(z)-\min_z u_t(z)$ and effect
magnitudes by workflow family.
Each interval spans task conditions within that family. $A_t$ is the mean
absolute conditional effect used to normalize recovery. The released
\path{results/conditional_effects.csv} preserves every source, mask pair, and
native outcome; the source inventory links each condition to its dataset or
record.
\begin{center}
{%
\fontsize{9}{11}\selectfont%
\setlength{\tabcolsep}{3pt}%
\begin{tabularx}{\linewidth}{@{}lRRR@{}}
\toprule
\rowcolor{WWBHeader}
\textbf{Workflow} & \textbf{Conditions} & \textbf{Response range ($\max-\min$)} & \textbf{Mean effect $A_t$} \\
\midrule
Regression & 5 & 0.0402--0.3847 & 0.0096--0.1018 \\
Retrieval & 3 & 0.3094--0.6394 & 0.0755--0.0992 \\
Image restoration & 5 & 0.1928--0.3306 & 0.0401--0.0678 \\
Classification & 5 & 0.0364--0.2799 & 0.0081--0.0391 \\
Forecasting & 5 & 0.0109--0.1594 & 0.0026--0.0384 \\
Clustering & 5 & 0.1733--0.8203 & 0.0433--0.1362 \\
Beat detection & 4 & 0.4464--0.6268 & 0.0538--0.1299 \\
Link prediction & 4 & 0.2470--0.4035 & 0.0199--0.0366 \\
\bottomrule
\end{tabularx}%
}

\end{center}

\section{Metric definitions and artifact grading}
\label{app:metrics}
\subsection{Coverage of joint interventions}
Let $x=z_0,z_1,\ldots,z_k=y$ be a path that switches each differing bit once.
Then
\begin{equation}
\sum_{j=1}^k[u(z_j)-u(z_{j-1})]=u(y)-u(x).
\end{equation}
Each summand is a signed conditional component effect. Exact effects therefore
determine every joint intervention and, with one anchor, every utility value.
Predicted effects derived from a complete table satisfy the same identity.
If every edge error is bounded by $\epsilon$, the triangle inequality bounds
the $k$-switch intervention error by $k\epsilon$. This statement applies to
maximum edge error; mean edge MAE describes average reconstruction quality.
Higher-order interactions remain represented through the background dependence
of conditional effects. Mean pair interactions provide a readable summary of
that structure and may average effects with different signs.

\subsection{Configuration choice and effect recovery}
Both two-bit tables below preserve the free anchors and select 10 as their
unique optimum for $0<\delta<1$.
\begin{center}\small
\begin{tabularx}{\linewidth}{@{}lRRRR@{}}
\toprule
\rowcolor{WWBHeader}
\textbf{Configuration} & \textbf{00} & \textbf{10} & \textbf{01} & \textbf{11} \\
\midrule
True utility & 0 & 1 & 0 & 0 \\
Predicted utility & 0 & 1 & $1-\delta$ & 0 \\
\bottomrule
\end{tabularx}
\end{center}
The mean absolute true edge effect is $A=1/2$ and edge MAE is
$E=(1-\delta)/2$, giving $R=\delta$. Thus exact choice can coexist with
arbitrarily weak effect recovery. Additional inactive bits embed the same
construction in four- and six-factor spaces. This example explains the
distinct evaluation targets; the main results measure their relationship
on native workflows.

\subsection{Native metrics and utility transformations}
Balanced accuracy is the arithmetic mean of class recalls. NMI uses
$2\operatorname{MI}(Y,C)/[H(Y)+H(C)]$, with contingency counts defining the
entropies and mutual information. Regression and forecasting use the arithmetic
mean absolute residual. Image restoration uses the square root of mean squared
pixel residual. Retrieval uses
\begin{equation}
\operatorname{NDCG@10}(q)=
\frac{\sum_{k=1}^{10} r_q(\pi_k)/\log_2(k+1)}
{\sum_{k=1}^{10}r_{q,(k)}/\log_2(k+1)},
\end{equation}
where $r_q$ is the original binary relevance and $r_{q,(k)}$ is its sorted ideal
gain. Utility averages this quantity over the fixed query cohort.

Event F1 is $2M/(P+T)$, where $P$ and $T$ are the numbers of interior predicted
and reference events, and $M$ is the maximum number of one-to-one pairs within
the matching tolerance. The sorted native matcher is checked by a separately
implemented maximum bipartite matcher. ROC AUC averages pairwise ranking credit over positive and negative examples, awarding one for a higher positive score and one-half for a tie.

Balanced accuracy, NMI, NDCG, event F1, and ROC AUC directly supply utility.
For native cost $c$ in regression, forecasting, or restoration,
\begin{equation}
u=\frac{1}{1+c/s},\qquad c=s\left(\frac{1}{u}-1\right).
\end{equation}
Regression uses the interquartile range of training targets for $s$.
Forecasting uses the interquartile range of the training series. Restoration
uses $s=0.1$. The fixed training statistics define the scale for every mask.

\subsection{Effect, interaction, and grid errors}
The main text defines conditional-effect MAE $E_t$ and recovery $R_t$. Pair
interaction $J_{ij}$ averages the second difference over all $2^{d-2}$ backgrounds,
and pair MAE is
\begin{equation}
I_t=\binom{d}{2}^{-1}\sum_{i<j}|\widehat J_{ij}-J_{ij}|.
\end{equation}
Grid MAE is $G_t=2^{-d}\sum_z|\hat u(z)-u(z)|$. The anonymous supplement
records $E_t$, $I_t$, $G_t$, and grid error on unobserved cells for every
scored artifact.

Selection chooses the largest predicted utility, then fewer enabled bits, then
the lexically smaller mask. Regret compares that mask's true utility with the
exhaustive optimum. Exact choice allows regret at most $10^{-12}$. A supplemental
near-optimal flag uses regret at most 0.02 utility, accompanied by an indicator
for tasks whose entire response range is below 0.02.

\subsection{Artifact validation and the strict diagnostic}
A complete artifact contains exactly one finite numerical utility in $[0,1]$
for each legal mask. Duplicate JSON keys, Boolean values in numerical cells,
nonfinite values, missing cells, and extra masks fail validation. The grader
replaces measured cells with their authoritative observed values before
computing effects and selection. Submission closes the episode.

Strict reconstruction is
\begin{equation}
S_t(\tau)=\mathbf{1}\!\left[
\max_{i,z:z_i=0}|\widehat\Delta_i(z)-\Delta_i(z)|
\leq\max\{10^{-12},\tau(\max u_t-\min u_t)\}\right].
\end{equation}
The development endpoint uses $\tau=0.1$. Missing delivery receives $R_t=0$,
$S_t=0$, and zero exact-choice credit. Its raw numerical errors remain undefined.
Shared estimators can still use the observations acquired before that failure.

\subsection{Measured effects and remaining reconstruction}
\label{app:calibration}
The complete-effect score includes knowledge supplied by direct measurement
and by predictions at unmeasured configurations. Let $K$ contain edges whose
endpoints are both observed, let $U$ contain the remaining edges, and let
$c=|K|/(|K|+|U|)$. Restoring authoritative observations makes the error on $K$
zero. Hence
\begin{equation}
E_t=(1-c)E_{U,t},\qquad
E_{U,t}=\frac{1}{|U|}\sum_{e\in U}|\widehat\Delta_e-\Delta_e|,\qquad
Q_{U,t}=E_{U,t}/A_t.
\end{equation}
Lower $Q_U$ indicates smaller error on effects involving unmeasured endpoints.
The task-wide $A_t$ fixes its scale across observation policies. Each policy determines the remaining edge set.

A simple calibration completes every unmeasured utility with the mean of that
episode's observed utilities. This baseline and the agent artifact receive
exactly the same observations and observed-cell restoration. The retrospective
analysis covers all \AgentTotalValid{} valid core artifacts; the table presents Flash
results separately for the original (O) and additional (A) cohorts. Repeated
executions are averaged within source and cohort. $N$ counts represented sources.
Columns labelled mean use the completion baseline; agent columns use delivered tables.
\begin{center}
{%
\fontsize{8}{10}\selectfont%
\setlength{\tabcolsep}{3pt}%
\begin{tabularx}{\linewidth}{@{}lRRRRRRRR@{}}
\toprule
\rowcolor{WWBHeader}
\textbf{Setting} & \textbf{$B$} & \textbf{Valid} & \textbf{$N$} & \textbf{Known edges} & \textbf{$R_{\rm mean}$} & \textbf{$R_{\rm agent}$} & \textbf{$Q_{U,\rm mean}$} & \textbf{$Q_{U,\rm agent}$} \\
\midrule
O, 4 factors & 4 & 6 & 6 & 10.4\% & 0.056 & 0.178 & 1.093 & 0.921 \\
O, 4 factors & 8 & 12 & 6 & 32.6\% & 0.287 & 0.632 & 1.054 & 0.542 \\
O, 6 factors, full & 20 & 6 & 6 & 6.8\% & 0.012 & 0.268 & 1.156 & 0.824 \\
O, 6 factors, opaque & 20 & 6 & 6 & 14.9\% & 0.011 & 0.257 & 1.183 & 0.899 \\
A, 4 factors & 4 & 12 & 6 & 9.9\% & 0.084 & 0.375 & 1.075 & 0.710 \\
A, 4 factors & 8 & 6 & 6 & 32.3\% & 0.213 & 0.621 & 1.156 & 0.567 \\
A, 6 factors, full & 20 & 10 & 6 & 8.6\% & 0.014 & 0.211 & 1.200 & 0.882 \\
A, 6 factors, opaque & 20 & 9 & 6 & 8.8\% & 0.028 & 0.283 & 1.172 & 0.790 \\
\bottomrule
\end{tabularx}%
}

\end{center}
For the original Flash cohort at $B=8$, 32.6\% of effects have two measured endpoints, leaving
67.4\% dependent on reconstruction. Mean completion gives recovery 0.287 and
$Q_U=1.054$, while the delivered tables give 0.632 and 0.542. Shared GP gives
recovery 0.698 and $Q_U=0.444$ on the same observations. The original six-factor views
also improve over mean completion, with 85.1\%--93.2\% of effects involving an
unmeasured endpoint. Per-episode coverage and errors remain in the anonymous supplement.
This analysis adds a calibration view of the saved artifacts and preserves
the original 4,206 numerical-control records and primary scoring definition.

\section{Numerical estimators and measurement policies}
\label{app:methods}
\subsection{Main-effect and pair-effect ridge}
For mask $z$, the feature vector contains an intercept and products of centered
bits $x_i=2z_i-1$. Degree one includes $x_i$; degree two adds $x_ix_j$ for all
$i<j$. For measured set $O$, ridge solves
\begin{equation}
\min_{\beta_0,\beta}\sum_{z\in O}
[u(z)-\beta_0-\phi(z)^\top\beta]^2+\alpha\|\beta\|_2^2.
\end{equation}
The intercept is unpenalized. Each candidate $\alpha$ in
$\{10^{-5},10^{-3},0.1,1,10\}$ is evaluated by leaving out one measured
observation at a time. Mean squared error selects the regularizer. Loss ties
at 14 decimal places prefer larger $\alpha$. The final table is clipped to
$[0,1]$ and measured cells are restored.

The sequential design policy uses the same factorial feature vector, including
the intercept, and selects the remaining mask with maximum
\begin{equation}
\phi_{\rm all}(z)^\top
\left(X_O^\top X_O+\operatorname{diag}(10^{-8},0.01,\ldots,0.01)\right)^{-1}
\phi_{\rm all}(z).
\end{equation}
Each chosen row enters the design matrix before the next selection. Utility
values enter ridge fitting; the design depends on feature geometry and the
already selected masks. Acquisition ties at ten decimal places prefer lexical
mask order. D1 denotes degree-one design and fitting, and D2 denotes degree-two
design and fitting. Rand2 samples a uniformly shuffled legal-mask sequence
without replacement and fits degree-two ridge. Each random setting has 20
recorded seed indices, numbered 0 through 19, added to seed base 20260907.

\subsection{Gaussian-process controls}
The all-order kernel is $k(z,z')=\exp[-h(z,z')/\ell]$, where $h$ is Hamming
distance. Ordinary kriging estimates a constant mean from measured values~\citep{gpml}.
For $C=K_O+\eta I$, let
\begin{equation}
q=C^{-1}\mathbf{1},\quad c=\mathbf{1}^\top q,\quad
P=C^{-1}-qq^\top/c.
\end{equation}
The leave-one-out residual vector has entries $(Py)_i/P_{ii}$. Its mean square
selects $\ell\in\{0.25,0.5,1,2,4,8\}$ and
$\eta\in\{10^{-6},10^{-3},0.03,0.3\}$. Loss ties at 14 decimal places prefer
larger length scale, then larger nugget. The nugget regularizes the fit to
deterministic utilities. Predictive means are clipped and observations restored.

Let $\Sigma$ be the ordinary-kriging covariance on the full grid and $D$ the
matrix taking all conditional component differences. The two acquisition
objectives for candidate $j$ apply predictive-variance reduction~\citep{cohnactive} to the grid and its linear contrasts, respectively, giving
\begin{equation}
a_G(j)=\frac{\|\Sigma_{:,j}\|_2^2}{\max(\Sigma_{jj}+\eta,10^{-12})},\qquad
a_E(j)=\frac{\|D\Sigma_{:,j}\|_2^2}{\max(\Sigma_{jj}+\eta,10^{-12})}.
\end{equation}
GP-G maximizes $a_G$ and GP-E maximizes $a_E$. Parameters are refitted after each
new observation. Acquisition ties at 12 decimal places prefer lexical mask
order. The grid and effect objectives share the same estimator.

\subsection{Program-equivalence controls}
\label{app:program_equiv_controls}
The six-factor control maps masks to representatives using the visible guards
in Section~\ref{app:options}. In regression, E resets to zero when C is zero,
and F resets to zero when B is zero. In clustering, E resets to zero when C
is zero. In restoration, E resets to zero when B is zero, and F resets to zero
when A is zero. In retrieval, C resets to zero when E is one. Classification
and forecasting retain all masks as separate representatives.

These rules yield 64, 36, 48, 64, 36, and 48 classes in classification,
regression, clustering, forecasting, restoration, and retrieval, respectively.
Every asserted alias matches the native prediction hash and utility of its
representative. CD1 and CD2 apply the corresponding ridge design and fit to
representatives. CGP-E computes its kernel on representative masks and uses
effect-variance acquisition. Purchased observations and inferred aliases are
recorded separately. Once every class is measured, the policy stops purchasing
and maps measured values to the complete grid. Thus actual credits $b$ can be
below the allowed budget $B$.

\subsection{Shared estimators and aggregation}
Shared fitting passes the exact observed set from an episode to D1, D2, or the
GP. It then evaluates the resulting complete table against the same reference.
The GP uses the fitting rule above with the observed set fixed. Its own
acquisition policy remains inactive during a shared fit. Episodes with missing
delivery retain their acquired observations for this analysis.

For random designs, metrics average over seeds within a source. Core
repeat studies average trials within a source. Family-macro scores then average
sources within each family and weight the participating families equally.
Four-factor controls use all 22 source instances, the paired six-factor
controls use the six paired
sources, and the added track uses four sources within each of its two families.
Agent--control comparisons select the same six source instances. Reported
counts retain their episode or source denominator explicitly.
The prospective typed cohort averages sources equally and conditions
$R_{\rm del}$ on valid delivery, with source-level resampling for its intervals.

\section{Core agent interface and execution environment}
\label{app:interface}
The core four-factor and paired six-factor studies expose four tools.
\texttt{current\_evidence()} returns the
public task, observations, remaining credits, calculation allowance, elapsed
time, remaining time, and episode status. \texttt{observe\_many(masks)} purchases
a batch of legal configurations. \texttt{calculate\_python(code)} runs numerical
Python with the current public task and observations.
The final submission tool registers a prediction table or a public estimator
selection. Appendix~\ref{app:typed_interface} specifies the independent
six-factor typed study.

The four-factor observation tool accepts 1--14 masks per call, and the paired
six-factor tool accepts 1--62. Repeated masks consume zero additional credits.
A batch is checked in full before charging or inserting observations. A valid
batch charges one credit per distinct newly measured mask. At $B$ purchased
measurements, the total observed set has at most $B+2$ cells. Native configuration
outcomes are cached from the reference construction.

Four-factor receipts include utility, native metric name, and native metric
value. Both views of the six-factor information study use utility-only receipts.
The full view contains workflow code, factor semantics, source information, and
input summaries. The opaque view contains a generic workflow identifier,
dimensions, factor order, legal masks, and the scalar utility contract. The
opaque view's identifiers and receipts preserve that information condition.

Every numerical call starts a fresh Python process with the current observations
and task. NumPy, SciPy, scikit-learn, and \texttt{wwb\_helpers} are available.
The process receives a 30-second CPU ceiling, 45-second wall ceiling, 2-billion-byte
address-space ceiling, and 2-million-byte file-size ceiling. CPU affinity uses
at most two available cores. The surrounding namespace mounts public task data,
helpers, and numerical dependencies. Network access and bundled scikit-learn
dataset labels are blocked. Printed outputs return to the agent for subsequent
reasoning. A maximum of 20 calculation calls applies to each episode.

The evaluated models are the recorded identifiers \texttt{deepseek-v4-flash}
and \texttt{deepseek-v4-pro}. Runs enable reasoning, request high effort, and
set an output-token cap of 32,768 with a 1,800-second outer session deadline.
The broker checks time before further investigation calls and accepts a final
submission while the outer session remains active. Tool receipts expose the
remaining investigation time. The prompt advises submission with 120 seconds
remaining. The public helper names are \texttt{ridge\_main} and
\texttt{ridge\_pair}, both recorded explicitly in submission metadata.

\section{Recorded agent prompts and task payloads}
\label{app:prompts}
The source bundle includes the complete frozen strings in
\path{prompts/four_system.txt}, \path{prompts/four_prompt.txt},
\path{prompts/six_system.txt}, and \path{prompts/six_prompt.txt}.
The anonymous supplement preserves the same templates in
\path{benchmark/prompts/}. Episode construction substitutes the
measurement budget, two anchor observations, and public task into the templates.

The instructions prioritize conditional-effect accuracy and require predictions
for all 16 or 64 configurations. They state the session and calculation limits,
describe the numerical helpers, and permit either a complete prediction table
or explicit helper selection. Submission is recommended with 120 seconds
remaining. Four-factor repetitions share the original prompt strings. Both
six-factor information views share the templates; their public task payloads
specify the information condition.

The structure diagnostic adds a first-call request for a JSON list of
inactive-option rules. Both arms receive this request; one additionally receives
verified rules in its task payload (Appendix~\ref{app:structure_probe}). Its exact
instruction and supplied payloads are included with the diagnostic records.
The task pack contains full workflow code and each agent's received payload.

\section{Study design and execution accounting}
\label{app:protocol}
\paragraph{Construction and analysis order.}
The 22-source four-factor inventory is specified before execution of its added
configurations. Paired construction extends the six initial agent-evaluation
sources, Spambase, Concrete, Wine, CO$_2$, Astronaut, and SciFact, with E and F
and requires agreement with their original subcubes. The signal and
graph extension lists eight sources and fixes its numerical methods and budgets
before native outcomes. Every listed source appears in the final catalog.
Acceptance uses native validity, cold replay, independent metric agreement,
and evaluation-boundary checks.

Plans for ridge/design, GP, and code-equivalence controls record source
selection, budgets, and code identities. The core agent cohort, paired
information study, original-cohort second $B=8$ Flash trial, and 66-episode repeat retain
their prompts, tools, deadlines, model identifiers, and execution dates.
The six-factor information study specifies shared-GP evaluation in its plan;
the core-cohort shared-GP fit is retrospective on saved observations.
The repeat comprises 42 four-factor and 24 six-factor information executions.
Tolerance, native-unit, and sign analyses use saved references and predictions.

\paragraph{Budgets and records.}
Four-factor controls use $B\in\{0,4,8,14\}$; paired six-factor controls use
$B\in\{0,4,8,14,20,32,62\}$; the eight added sources use
$B\in\{0,8,20,32,62\}$. The 4,206 records comprise 1,936 four-factor ridge/design
runs, 924 paired ridge/design runs, 84 code-equivalence ridge runs, 302 GP runs,
and 960 added-source runs. Random designs use 20 seed indices. Full-observation
endpoints purchase every remaining mask or program-equivalence class.

\begin{center}\small
\begin{tabularx}{\linewidth}{@{}lRRRR@{}}
\toprule
\rowcolor{WWBHeader}
\textbf{Agent study} & \textbf{$d$} & \textbf{$B$} & \textbf{Episodes} & \textbf{Delivered} \\
\midrule
Flash and Pro original cohort & 4 & 4, 8 & 24 & 20 \\
Flash full/opaque information & 6 & 20 & 12 & 12 \\
Flash original, second $B=8$ trial & 4 & 8 & 6 & 6 \\
Additional Flash and Pro & 4 & 4, 8 & 42 & 37 \\
Additional Flash information & 6 & 20 & 24 & \AdditionalInformationValid{} \\
Structure diagnostic & 6 & 20 & 8 & 8 \\
Typed estimator-menu study & 6 & 20 & 8 & 8 \\
Fixed-observation inference & 6 & 0 & 4 & 4 \\
Typed source cohort & 4 & 8 & 24 & 23 \\
Added beat/graph workflows & 6 & 32 & 8 & 6 \\
\midrule
\textbf{All studies} & & & \textbf{160} & \textbf{143} \\
\bottomrule
\end{tabularx}
\end{center}

The first five rows comprise 108 core episodes. The added-family row includes
six delivered tables and two source attempts without final artifacts; the
delivered-table analysis in Appendix~\ref{app:valid_added} uses all six.
Core accounting retains every attempted provider request and excludes five
zero-request administrative records. Four original-cohort Pro episodes end at
their 1,800-second deadlines after 4, 12, 23, and 14 HTTP 429 responses;
the original four-factor cohort records 75 HTTP 429 responses in total.
Five additional-cohort Pro episodes also end without final artifacts, with
zero HTTP 429 responses across that cohort.
The typed-source Pro Astronaut episode runs in the four-session final batch,
purchases all eight measurements, and reaches its 1,800-second deadline
without a final table. The original, information, and repeat studies record
748, 304, and 155 terminal requests, respectively. Provider identifiers,
response status, usage, timing, and observation events remain in the
anonymous supplement.

\section{Episode results and paired comparisons}
\label{app:agentresults}
Table~\ref{tab:models} separates delivery, all-attempt recovery, recovery
conditional on delivery, and shared-GP recovery on acquired observations.

\begin{table}[!ht]
\caption{Four-factor agent evaluation on six matched sources. Delivery is
complete artifacts over attempts. $R_{\rm all}$ averages all attempts with
zero for missing artifacts; $R_{\rm del}$ averages delivered artifacts;
$R_{\rm GP}$ fits a common GP to observations from all attempts. Exact and
strict count episodes.}
\vspace{4pt}
\centering\small
{%
\fontsize{8}{10}\selectfont%
\setlength{\tabcolsep}{3pt}%
\begin{tabularx}{\linewidth}{@{}llRRRRRRR@{}}
\toprule
\rowcolor{WWBHeader}
\textbf{Model} & \textbf{Cohort} & \textbf{$B$} & \textbf{Delivery} & \textbf{$R_{\rm all}$} & \textbf{$R_{\rm del}$} & \textbf{$R_{\rm GP}$} & \textbf{Exact} & \textbf{Strict} \\
\midrule
Flash & Original & 4 & 6/6 & 0.178 & 0.178 & 0.226 & 2 & 0 \\
Flash & Original & 8 & 12/12 & 0.632 & 0.632 & 0.698 & 5 & 1 \\
Pro & Original & 4 & 4/6 & 0.043 & 0.064 & 0.346 & 2 & 0 \\
Pro & Original & 8 & 4/6 & 0.366 & 0.549 & 0.692 & 3 & 0 \\
Flash & Additional & 4 & 12/12 & 0.375 & 0.375 & 0.338 & 5 & 0 \\
Flash & Additional & 8 & 6/6 & 0.621 & 0.621 & 0.720 & 2 & 0 \\
Pro & Additional & 4 & 10/12 & 0.308 & 0.370 & 0.315 & 2 & 0 \\
Pro & Additional & 8 & 9/12 & 0.503 & 0.671 & 0.667 & 6 & 1 \\
\bottomrule
\end{tabularx}%
}

\label{tab:models}
\end{table}

The compact calibration table retains repetition variability and a mean-fill
reference. Within-source SD averages the sample SD of all-attempt recovery
across repetitions within each source; a dash marks one execution per source.
Mean-fill recovery preserves purchased scores and fills other configurations
with their observed mean. Its denominator $n_{\rm fill}$ counts delivered
tables. The released \path{results/agent_index.json} retains all
\AgentTotalEpisodes{} core episode records, predictions, measurements, helper
choices, and execution metadata.
\begin{center}
{%
\fontsize{8}{10}\selectfont%
\setlength{\tabcolsep}{3pt}%
\begin{tabularx}{\linewidth}{@{}llRRRR@{}}
\toprule
\rowcolor{WWBHeader}
\textbf{Model} & \textbf{Cohort} & \textbf{$B$} & \textbf{Within-source SD} & \textbf{Mean-fill $R$} & \textbf{$n_{\rm fill}$} \\
\midrule
Flash & Original & 4 & \textemdash{} & 0.056 & 6 \\
Flash & Original & 8 & 0.073 & 0.287 & 12 \\
Pro & Original & 4 & \textemdash{} & 0.045 & 4 \\
Pro & Original & 8 & \textemdash{} & 0.326 & 4 \\
Flash & Additional & 4 & 0.105 & 0.084 & 12 \\
Flash & Additional & 8 & \textemdash{} & 0.213 & 6 \\
Pro & Additional & 4 & 0.130 & 0.115 & 10 \\
Pro & Additional & 8 & 0.080 & 0.273 & 9 \\
\bottomrule
\end{tabularx}%
}

\end{center}

\subsection{Information supplied to the agent}
\label{app:context}
Full and opaque conditions share sources, reference tables, budgets, tools,
helpers, utility-only receipts, and deadlines. Full information adds code,
option semantics, source information, and input summaries jointly. Each source
has one attempted pair in the original cohort and two in the additional cohort.
The table reports $R_{\rm all}$ and shared-GP recovery;
$\Delta$ is full minus opaque.
\begin{center}
{%
\fontsize{8}{10}\selectfont%
\setlength{\tabcolsep}{3pt}%
\begin{tabularx}{\linewidth}{@{}llRRRRRRR@{}}
\toprule
\rowcolor{WWBHeader}
\textbf{Cohort} & \textbf{Source} & \textbf{Valid pairs} & \multicolumn{3}{c}{\bfseries $R_{\rm all}$} & \multicolumn{3}{c}{\bfseries Shared GP $R$} \\
\cmidrule(lr){4-6}\cmidrule(lr){7-9}
 & & & Full & Opaque & $\Delta$ & Full & Opaque & $\Delta$ \\
\midrule
O & astronaut & 1/1 & 0.294 & 0.255 & +0.039 & 0.379 & 0.137 & +0.242 \\
O & co2 & 1/1 & 0.244 & 0.049 & +0.196 & 0.263 & 0.131 & +0.133 \\
O & concrete & 1/1 & 0.298 & 0.642 & -0.343 & 0.796 & 0.847 & -0.051 \\
O & scifact & 1/1 & 0.522 & 0.407 & +0.115 & 0.598 & 0.468 & +0.130 \\
O & spambase & 1/1 & 0.000 & 0.000 & +0.000 & 0.000 & 0.389 & -0.389 \\
O & wine & 1/1 & 0.247 & 0.191 & +0.056 & 0.205 & 0.414 & -0.209 \\
\midrule
A & astronaut & 1/2 & 0.275 & 0.000 & +0.275 & 0.217 & 0.112 & +0.105 \\
A & co2 & 1/2 & 0.129 & 0.102 & +0.027 & 0.262 & 0.161 & +0.101 \\
A & concrete & 2/2 & 0.505 & 0.754 & -0.249 & 0.771 & 0.843 & -0.072 \\
A & scifact & 1/2 & 0.153 & 0.357 & -0.204 & 0.300 & 0.496 & -0.195 \\
A & spambase & 2/2 & 0.042 & 0.000 & +0.042 & 0.039 & 0.203 & -0.164 \\
A & wine & 1/2 & 0.006 & 0.193 & -0.187 & 0.002 & 0.160 & -0.158 \\
\bottomrule
\end{tabularx}%
}

\end{center}
Cohorts O and A denote original and additional executions. Valid pairs count
pairs with two delivered artifacts over all attempted pairs. Means retain all
attempts, including $R_{\rm all}=0$ after a missing submission.
The original mean $R_{\rm all}$ difference is 0.0103 and its shared-GP difference
is $-0.0242$. Additional-cohort differences are $-0.0491$ for $R_{\rm all}$
and $-0.0638$ for shared GP. The eight additional pairs with two delivered
artifacts cover all six sources; their source-macro differences are $-0.1198$
and $-0.0914$, respectively. These conditional summaries describe the paired
deliveries, while the main comparison retains every attempted pair.

\subsection{Repeated high-budget agent runs}
\label{app:repeatresults}
Both original Flash $B=8$ trials and the additional execution are retained for
each source. The table keeps their cohorts separate. Parentheses give the
sample standard deviation where repetitions are available; a dash denotes
a single execution. The final column reports matched D2 recovery.
\begin{center}
{%
\fontsize{8}{10}\selectfont%
\setlength{\tabcolsep}{3pt}%
\begin{tabularx}{\linewidth}{@{}lRRRRR@{}}
\toprule
\rowcolor{WWBHeader}
\textbf{Source} & \multicolumn{2}{c}{\bfseries Flash, original} & \multicolumn{2}{c}{\bfseries Flash, additional} & \textbf{Designed D2} \\
\cmidrule(lr){2-3}\cmidrule(lr){4-5}
 & $R_{\rm all}$ (SD) & Shared GP & $R_{\rm all}$ (SD) & Shared GP & $R$ \\
\midrule
astronaut & 0.356 (0.144) & 0.645 & 0.270 (\textemdash{}) & 0.523 & 0.634 \\
co2 & 0.365 (0.026) & 0.505 & 0.569 (\textemdash{}) & 0.667 & 0.400 \\
concrete & 0.853 (0.011) & 0.877 & 0.842 (\textemdash{}) & 0.913 & 0.888 \\
scifact & 0.803 (0.096) & 0.799 & 0.762 (\textemdash{}) & 0.761 & 0.799 \\
spambase & 0.637 (0.066) & 0.667 & 0.684 (\textemdash{}) & 0.683 & 0.508 \\
wine & 0.778 (0.095) & 0.691 & 0.599 (\textemdash{}) & 0.775 & 0.533 \\
\bottomrule
\end{tabularx}%
}

\end{center}
Across the 12 original episodes, five select an exact optimum and one passes
strict reconstruction. Original mean recovery is 0.6321 and shared GP recovery
is 0.6976. The six additional episodes have mean recovery 0.6211 and shared GP
recovery 0.7205, with two exact choices and zero strict reconstructions.
The anonymous supplement indexes every measurement batch and prediction.

\subsection{Rule recognition and application}
\label{app:structure_probe}
This rule-reporting diagnostic uses all four six-factor workflows with nontrivial
code equivalences. Each source receives one full-code and one code-plus-rules
Flash episode at $B=20$, with arm order alternated across sources. Both arms
use the full task and the same first-call instruction to report inactive-option
rules before buying measurements. The report counts toward the 20 calculation
calls and receives no correctness feedback. The rules arm additionally receives
six registered guard relations across the four sources. Independent checks
confirm identical native predictions for all 96 registered equivalent endpoint
pairs. Models, tools, anchors, and session limits match the information study.
These eight episodes were run on September 22, 2026 (UTC), as a separate
cohort with an elicited reporting task.

The table separates observable rule reporting, acquisition, and prediction.
Rules counts correctly reported registered relations. Redundant counts purchases
within an equivalence class already represented by an anchor or paid measurement.
Equiv.\ error is the mean absolute predicted-utility difference across the
registered equivalent pairs, divided by the source's utility range. Recovery
retains all attempted episodes. The final column is an offline class-consistent
control applied to each frozen submission. A class containing a
measured mask receives that observation; other classes receive the mean of their
submitted predictions. The projection preserves every acquired observation and
uses no new measurements. All eight episodes deliver valid artifacts.
\begin{center}
{\small
\begin{tabularx}{\linewidth}{@{}llRRRRR@{}}
\toprule
\rowcolor{WWBHeader}
\textbf{Source} & \textbf{Information} & \textbf{Rules} & \textbf{Redundant} & \textbf{Equiv. error} & \textbf{Recovery} & \textbf{Class $R$} \\
\midrule
concrete & Code & 2/2 & 0/20 & 4.88\% & 0.680 & 0.823 \\
concrete & Code + rules & 2/2 & 0/20 & 10.21\% & 0.646 & 0.748 \\
\midrule
wine & Code & 1/1 & 0/20 & 16.97\% & 0.000 & 0.000 \\
wine & Code + rules & 1/1 & 0/20 & 15.54\% & 0.213 & 0.290 \\
\midrule
astronaut & Code & 2/2 & 2/20 & 16.77\% & 0.000 & 0.415 \\
astronaut & Code + rules & 2/2 & 0/20 & 12.14\% & 0.334 & 0.745 \\
\midrule
scifact & Code & 1/1 & 0/20 & 8.89\% & 0.308 & 0.485 \\
scifact & Code + rules & 1/1 & 0/20 & 1.39\% & 0.524 & 0.549 \\
\bottomrule
\end{tabularx}
}

\end{center}

Both arms correctly report all six registered rules. The rules-supplied arm
has higher recovery on three sources and lower recovery on concrete; arm means are 0.247 and
0.429. Shared GP reaches 0.382 and 0.471. The code arm spends 2/80 credits
on equivalent repeats and the rules arm spends none. All eight final artifacts
use public ridge helpers and violate equivalences, with mean equivalence
errors of 11.88\% and 9.82\% of source utility range. The class-consistent
control scores mean recovery 0.507 versus 0.338 for the delivered tables at
fixed observations. It has lower raw effect error in all eight episodes and
higher recovery in seven; the strict score remains zero. These source pairs
separate reported rules from delivered ridge-helper predictions and quantify
the value of registered equivalences in frozen prediction tables.
The anonymous supplement includes the episode records and independent
regrading under \path{results/structure/}.

\subsection{Typed submissions and fixed-observation inference}
\label{app:typed_interface}
\paragraph{Submission format.}
Both matched arms accept complete 64-score prediction objects and record
the submitted table as the final artifact. The estimator menu additionally
exposes GP fitting and agent-supplied structure rules.

\paragraph{Matched typed-interface episodes.}
The matched typed and fixed-observation studies were executed on September
25, 2026 (UTC), using API identifier \texttt{deepseek-v4-flash}.
The prospective study pairs a ridge-only submission interface with a ridge,
GP, and agent-rule menu on concrete, wine, astronaut, and SciFact at $B=20$.
Both arms use Flash, the same code-visible task and typed complete-table
argument, two free anchors, 20 purchased outcomes, 20 calculation calls, and
a 1,800-second deadline. Arm order alternates across sources. Both prompts
request a first-call report of inactive-option rules. The menu also exposes
\texttt{fit\_estimator} in numerical Python. Three of four menu
trajectories successfully call the rule-aware GP helper during calculation.
All eight final calls submit direct tables without estimator or rule
parameters.

Both arms deliver 4/4 artifacts and report 6/6 registered rules. Seven of
eight tables satisfy every verified equivalence. The paired menu-minus-ridge
mean is $-0.0923$, with one positive and three negative source contrasts.
Source-level results appear below.
\begin{center}\small
\begin{tabularx}{\linewidth}{@{}lRRRR@{}}
\toprule
\rowcolor{WWBHeader}
\textbf{Source} & \textbf{Ridge $R$} & \textbf{Menu $R$} & \textbf{Menu $-$ ridge} & \textbf{Clean tables} \\
\midrule
Concrete & 0.8990 & 0.9291 & $+0.0301$ & 2/2 \\
Wine & 0.4465 & 0.4455 & $-0.0010$ & 2/2 \\
Astronaut & 0.8865 & 0.6758 & $-0.2107$ & 1/2 \\
SciFact & 0.6071 & 0.4194 & $-0.1877$ & 2/2 \\
\bottomrule
\end{tabularx}
\end{center}
Offline common-estimator fits hold each episode's 22 observations fixed.
Registered-rule columns use the verified code rules from the frozen plan.
\begin{center}\small
\begin{tabularx}{\linewidth}{@{}lRRRR@{}}
\toprule
\rowcolor{WWBHeader}
\textbf{Acquisition arm} & \textbf{Shared D2} & \textbf{D2 + rules} & \textbf{Shared GP} & \textbf{GP + rules} \\
\midrule
Ridge-only & 0.319 & 0.601 & 0.509 & 0.666 \\
Menu & 0.224 & 0.492 & 0.350 & 0.535 \\
\bottomrule
\end{tabularx}
\end{center}

\paragraph{Fresh inference from fixed observations.}
Each fresh Flash agent receives the 22 observations from the matched typed
ridge-only episode, with zero remaining measurement credits and a 900-second
session. The final call requires a named estimator and agent-supplied rules.
All four agents report and submit their source's registered rules, covering
six registered source-specific relations across the study. Three choose GP and
one chooses main-effect ridge. All 96 checked native-equivalent pairs receive
identical predictions. The four final tables attain mean recovery 0.682;
the separate source-matched typed ridge-only direct tables attain 0.710.
\begin{center}\small
\begin{tabularx}{\linewidth}{@{}lRRRR@{}}
\toprule
\rowcolor{WWBHeader}
\textbf{Source} & \textbf{Typed direct $R$} & \textbf{Fixed $R$} & \textbf{Estimator} & \textbf{Rules / violations} \\
\midrule
Concrete & 0.899 & 0.928 & GP & 2/2, 0/32 \\
Wine & 0.446 & 0.407 & GP & 1/1, 0/16 \\
Astronaut & 0.887 & 0.767 & D1 & 2/2, 0/32 \\
SciFact & 0.607 & 0.625 & GP & 1/1, 0/16 \\
\bottomrule
\end{tabularx}
\end{center}
The anonymous supplement includes both frozen plans, the typed submissions,
intermediate calculations, and final regrading.

\subsection{Prospective typed source cohort}
\label{app:typed_sources}
The frozen study pairs Flash and Pro on 12 four-factor sources at $B=8$ under
the same typed task and tool interface. Six sources are the paired four/six-factor
agent sources; the plan adds the alphabetically first agent-unevaluated source
ID in each family before this cohort's model outcomes. Every episode receives two free anchors, up to
eight purchased outcomes, 20 numerical calculation calls, and a 1,800-second
deadline. The paired source is the aggregation unit. All 24 episodes purchase
eight outcomes, and 23 deliver complete tables. The Pro Astronaut episode
reaches its session deadline after its eighth purchase without a final table.
The first 20 jobs follow the frozen paired launch order. A recorded scheduling
amendment runs the final Astronaut and SciFact jobs across sources with four
concurrent sessions; tasks, prompts, budgets, estimators, and scoring remain
fixed. The amendment is included with the anonymous supplement.
The 24 episodes were executed on September 25, 2026 (UTC), using API
identifiers \texttt{deepseek-v4-flash} and \texttt{deepseek-v4-pro}.

Table~\ref{tab:typed_sources_main} summarizes source groups, delivery, and
inference in the main text. $R_{\rm all}$ assigns zero recovery to a missing final artifact;
$R_{\rm del}$ averages delivered artifacts; $R_{\rm GP}$ fits a common GP to
the same purchased masks and utilities in every episode. The paired gain is
$R_{\rm all}-R_{\rm GP}$, with descriptive 2.5th--97.5th percentile
source-bootstrap ranges from 20,000
resamples. The GP uses purchased mask--utility pairs; agent tables may also
use public workflow code and sandbox calculations.

Flash submits a direct typed prediction table on all 12 sources. Pro submits
ten direct tables, selects the named GP on Coffee, and has the one Astronaut
non-delivery. The source-level table preserves every attempt. Flash delivers
on all 12 sources; Pro delivery is shown explicitly for each source. The
Astronaut Pro GP value uses its purchased observations while its
$R_{\rm all}$ is zero. D2 GP fits the observations from the frozen
pair-effect design for that source.
\begin{center}
{\fontsize{8}{10}\selectfont\setlength{\tabcolsep}{2.5pt}%
\begin{tabularx}{\linewidth}{@{}llRRRRRR@{}}
\toprule
\rowcolor{WWBHeader}
\textbf{Source} & \textbf{Group} & \textbf{Flash $R$} & \textbf{Flash GP} & \textbf{Pro valid} & \textbf{Pro $R_{\rm all}$} & \textbf{Pro GP} & \textbf{D2 GP} \\
\midrule
Banknote & New & 0.829 & 0.797 & 1 & 0.768 & 0.623 & 0.745 \\
Airfoil & New & 0.801 & 0.695 & 1 & 0.718 & 0.691 & 0.717 \\
Dry Bean & New & 0.852 & 0.637 & 1 & 0.853 & 0.590 & 0.590 \\
El Ni\~no & New & 0.987 & 0.960 & 1 & 0.992 & 0.981 & 0.967 \\
Coffee & New & 0.881 & 0.538 & 1 & 0.732 & 0.732 & 0.732 \\
ArguAna & New & 0.861 & 0.689 & 1 & 0.688 & 0.689 & 0.709 \\
\midrule
Spambase & Original & 0.562 & 0.666 & 1 & 0.652 & 0.514 & 0.560 \\
Concrete & Original & 0.922 & 0.885 & 1 & 0.872 & 0.872 & 0.885 \\
Wine & Original & 0.895 & 0.713 & 1 & 0.901 & 0.418 & 0.494 \\
CO$_2$ & Original & 0.499 & 0.437 & 1 & 0.245 & 0.667 & 0.667 \\
Astronaut & Original & 0.894 & 0.259 & 0 & 0.000 & 0.707 & 0.584 \\
SciFact & Original & 0.811 & 0.747 & 1 & 0.871 & 0.821 & 0.821 \\
\bottomrule
\end{tabularx}}
\end{center}

Across Pro's 11 delivered sources, $R_{\rm del}=0.754$ and the GP mean on
those same sources is 0.691, giving a delivered-only difference of $+0.063$.
The named GP submission on Coffee gives an exact zero difference.
The shared-GP acquisition contrast compares agent-purchased observations
with the frozen D2 design. On the six new sources its mean is $-0.024$ for
Flash (descriptive source-bootstrap range $[-0.096,0.029]$) and $-0.026$ for Pro
($[-0.066,0.002]$). Across all 12 sources the respective means are $-0.037$
($[-0.121,0.041]$) and $-0.014$ ($[-0.044,0.019]$). On the six paired sources,
the Flash mean is $-0.051$, with two positive, one tied, and three negative
source contrasts. A post-hoc sensitivity
selects the strongest of the same-observation main-effect ridge, pair-effect
ridge, and GP fits separately for each source. Flash delivered tables exceed
this per-source best fit on all six new sources by 0.114 mean recovery and on
10 of 12 sources overall by 0.095. Pro exceeds it on four new sources, with
a six-source mean difference of 0.073. The frozen episode plan, amendment,
all 24 attempts, independent regrading, and source-bootstrap analysis appear
in the anonymous supplement.

\clearpage
\section{Sensitivity and additional-workflow results}
\label{app:sensitivity}
\subsection{Sensitivity to individual sources}
\label{app:source_sensitivity}
Each comparison is paired by source. The original Flash comparison first
averages its two trials within each source; the additional comparison uses its
separate execution on each source. We report the mean and median recovery change,
counts of positive, tied, and negative sources, and the range of paired means
after omitting each of the six sources in turn. These ranges describe the
composition of the evaluated catalog.
\begin{figure}[!ht]
\centering
\includegraphics[width=\linewidth]{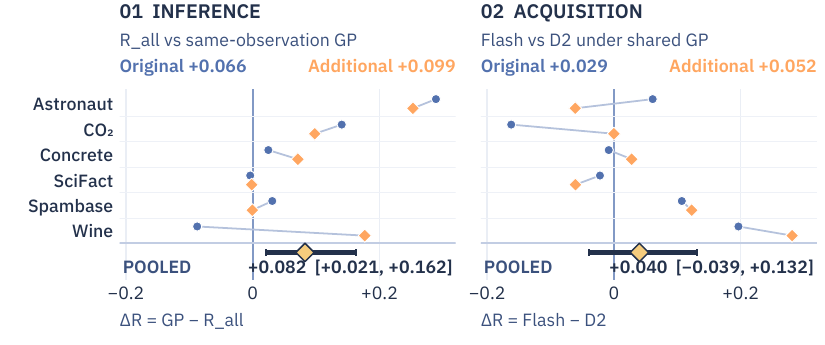}
\caption{Source-level acquisition and inference contrasts at $B=8$. Each
source has one original-cohort point, averaging two Flash episodes, and one
additional-cohort point from one episode. The left panel compares Flash
$R_{\rm all}$ with a GP fitted to the same observations. The right panel
compares Flash and D2 observations under that GP. Pooled means average cohorts
within source; bars are descriptive source-bootstrap percentile ranges from
20,000 resamples.}
\label{fig:acquisition_inference}
\end{figure}
\begin{center}
{%
\fontsize{9}{11}\selectfont%
\setlength{\tabcolsep}{3pt}%
\begin{tabularx}{\linewidth}{@{}lRRRRRR@{}}
\toprule
\rowcolor{WWBHeader}
\textbf{Comparison} & \textbf{$B$} & \textbf{Mean $\Delta R$} & \textbf{Median} & \textbf{$+ / = / -$} & \multicolumn{2}{c}{\bfseries Source omission range} \\
\midrule
Shared GP, Flash original & 8 & +0.066 & +0.028 & 4 / 0 / 2 & +0.021 & +0.096 \\
Shared GP, Flash additional & 8 & +0.099 & +0.085 & 4 / 0 / 2 & +0.069 & +0.120 \\
Code GP-E over GP-E & 20 & +0.214 & +0.166 & 4 / 2 / 0 & +0.121 & +0.257 \\
Code D2 over D2 & 20 & +0.233 & +0.231 & 4 / 2 / 0 & +0.167 & +0.279 \\
Code GP-E over GP-E & 32 & +0.159 & +0.131 & 4 / 2 / 0 & +0.096 & +0.191 \\
Code D2 over D2 & 32 & +0.205 & +0.136 & 4 / 2 / 0 & +0.113 & +0.246 \\
\bottomrule
\end{tabularx}%
}

\end{center}
\paragraph{Paired source contrasts.}
At $B=8$, each contrast fits the same GP to observations from two acquisition
policies or compares that fit with agent $R_{\rm all}$. Each of the six
sources is a resampling cluster. The original cohort first averages its two
episodes within source; the additional cohort has one episode per source.
Pooled contrasts average the two cohort contrasts within source, retaining six
source-level resampling units. Descriptive percentile ranges use 20,000 source-bootstrap
resamples. The final column counts positive, tied, and negative sources.
\begin{center}
{%
\fontsize{8}{10}\selectfont%
\setlength{\tabcolsep}{3pt}%
\begin{tabularx}{\linewidth}{@{}llRRR@{}}
\toprule
\rowcolor{WWBHeader}
\textbf{Flash cohort} & \textbf{Paired contrast} & \textbf{Mean $\Delta R$} & \textbf{Source bootstrap range} & \textbf{$+ / = / -$} \\
\midrule
Original & GP $-R_{\rm all}$ & +0.066 & $[-0.021, +0.167]$ & 4/0/2 \\
Original & GP $-$ D2 & +0.029 & $[-0.064, +0.115]$ & 3/0/3 \\
Original & GP $-$ GP-E & +0.058 & $[-0.046, +0.164]$ & 4/0/2 \\
\midrule
Additional & GP $-R_{\rm all}$ & +0.099 & $[+0.028, +0.172]$ & 4/0/2 \\
Additional & GP $-$ D2 & +0.052 & $[-0.031, +0.155]$ & 3/1/2 \\
Additional & GP $-$ GP-E & +0.081 & $[-0.023, +0.202]$ & 3/1/2 \\
\midrule
Pooled & GP $-R_{\rm all}$ & +0.082 & $[+0.021, +0.162]$ & 5/0/1 \\
Pooled & GP $-$ D2 & +0.040 & $[-0.039, +0.132]$ & 4/0/2 \\
Pooled & GP $-$ GP-E & +0.069 & $[-0.029, +0.180]$ & 3/0/3 \\
\bottomrule
\end{tabularx}%
}

\end{center}
The pooled Flash gain from shared GP over delivery is $+0.082$, with a
descriptive source-bootstrap range of $[+0.021,+0.162]$. GP fitted to Flash
observations also has higher six-source means than GP fitted to D2 or GP-E
observations in both cohorts. Their source-bootstrap ranges show variation across
acquisition policies.
The original shared-GP mean advantage ranges from 0.021 to 0.096 under source omission.
The additional Flash cohort has a mean gain of 0.099, with a source-omission
range of 0.069 to 0.120. Four sources improve and two decline in each cohort.
Program-equivalence controls improve on all four sources with nontrivial
classes. Classification and forecasting retain identity maps and yield the
same predictions as their corresponding controls. At both reported budgets,
every source-omission mean remains positive. The anonymous supplement includes
all per-source contrasts and source-omission means.

\subsection{Strict-error tolerances}
Each row summarizes five tolerances on the same saved predictions. Track four
uses all 22 four-factor sources; track six uses the six paired sources. Strict
counts use maximum conditional-effect error relative to the response range.
Exact is the number of optimal configuration choices, and the final column
averages recovery among those exact-choice cases.
\begin{center}
{%
\fontsize{8}{10}\selectfont%
\setlength{\tabcolsep}{3pt}%
\begin{tabularx}{\linewidth}{@{}llRRRRRRRRR@{}}
\toprule
\rowcolor{WWBHeader}
\textbf{Track} & \textbf{Method} & \textbf{$B$} & \textbf{$N$} & \multicolumn{5}{c}{\bfseries Strict count at tolerance $\tau$} & \textbf{Exact} & \textbf{$R\mid\mathrm{Exact}$} \\
\cmidrule(lr){5-9}
 & & & & .025 & .05 & .10 & .20 & .40 & & \\
\midrule
4 factors & D1 & 8 & 22 & 0 & 1 & 1 & 5 & 13 & 15 & 0.618 \\
4 factors & D2 & 8 & 22 & 1 & 1 & 3 & 8 & 20 & 15 & 0.634 \\
4 factors & GP-G & 8 & 22 & 1 & 1 & 2 & 10 & 19 & 15 & 0.669 \\
4 factors & GP-E & 8 & 22 & 1 & 1 & 1 & 12 & 19 & 16 & 0.717 \\
\midrule
6 factors & D1 & 8 & 6 & 0 & 0 & 0 & 0 & 1 & 2 & 0.405 \\
6 factors & D2 & 8 & 6 & 0 & 0 & 0 & 0 & 0 & 0 & \textemdash{} \\
6 factors & GP-G & 8 & 6 & 0 & 0 & 0 & 1 & 1 & 0 & \textemdash{} \\
6 factors & GP-E & 8 & 6 & 0 & 0 & 0 & 0 & 2 & 0 & \textemdash{} \\
6 factors & Code D1 & 8 & 6 & 0 & 0 & 0 & 1 & 1 & 2 & 0.063 \\
6 factors & Code D2 & 8 & 6 & 0 & 0 & 0 & 0 & 0 & 1 & 0.000 \\
6 factors & Code GP-E & 8 & 6 & 0 & 0 & 0 & 1 & 1 & 2 & 0.544 \\
\midrule
6 factors & D1 & 32 & 6 & 0 & 0 & 0 & 1 & 2 & 2 & 0.431 \\
6 factors & D2 & 32 & 6 & 0 & 0 & 1 & 1 & 2 & 2 & 0.598 \\
6 factors & GP-G & 32 & 6 & 0 & 0 & 1 & 1 & 4 & 2 & 0.682 \\
6 factors & GP-E & 32 & 6 & 0 & 0 & 1 & 1 & 4 & 2 & 0.704 \\
6 factors & Code D1 & 32 & 6 & 0 & 2 & 2 & 3 & 3 & 3 & 0.786 \\
6 factors & Code D2 & 32 & 6 & 1 & 1 & 3 & 3 & 3 & 2 & 0.987 \\
6 factors & Code GP-E & 32 & 6 & 1 & 2 & 2 & 3 & 4 & 3 & 0.925 \\
\bottomrule
\end{tabularx}%
}

\end{center}
This profile connects the continuous metric to several maximum-error criteria.
For example, D2 at $B=8$ passes on 3/22 sources at the 10\% tolerance and
8/22 at 20\%, while its 15 exact choices remain the same. Native-case response
scales appear in Appendix~\ref{app:conditions}.

\subsection{Magnitude of conditional reversals}
A reversal requires at least one positive and one negative background effect
whose magnitudes both exceed the indicated floor. Source counts merge paired
dimensional variants. The two largest floors identify substantial conditional
changes within the fixed response range.
\begin{center}\small
\begin{tabularx}{\linewidth}{@{}lRRR@{}}
\toprule
\rowcolor{WWBHeader}
\textbf{Effect floor} & \textbf{Components / 172} & \textbf{Conditions / 36} & \textbf{Sources / 30} \\
\midrule
$10^{-12}$ utility & 121 & 35 & 29 \\
1\% of response range & 98 & 33 & 28 \\
5\% of response range & 59 & 24 & 20 \\
10\% of response range & 31 & 16 & 14 \\
\bottomrule
\end{tabularx}
\end{center}

\subsection{Native-unit sensitivity}
\label{app:native_sensitivity}
The analysis applies the inverse utility map to saved predictions and computes
effects in native units. Fitted models and observed sets remain fixed.
All 3,246 analysed original control records have finite inverses. Raw native
errors are reported by source in the anonymous supplement, preserving each source's units.
The table compares family-macro recovery at the main four-factor budget and
both reported paired budgets. $N$ counts sources. Native strict counts successful records across all seeds, with 20 seeds for each random design.
\begin{center}
{%
\fontsize{9}{11}\selectfont%
\setlength{\tabcolsep}{3pt}%
\begin{tabularx}{\linewidth}{@{}llRRRRR@{}}
\toprule
\rowcolor{WWBHeader}
\textbf{Track} & \textbf{Method} & \textbf{$B$} & \textbf{$N$} & \textbf{Utility $R$} & \textbf{Native $R$} & \textbf{Native strict} \\
\midrule
4 factors & D1 & 8 & 22 & 0.571 & 0.562 & 1/22 \\
4 factors & D2 & 8 & 22 & 0.612 & 0.633 & 3/22 \\
4 factors & Random D2 & 8 & 22 & 0.503 & 0.505 & 14/440 \\
4 factors & GP-G & 8 & 22 & 0.661 & 0.669 & 2/22 \\
4 factors & GP-E & 8 & 22 & 0.701 & 0.712 & 2/22 \\
\midrule
6 factors & D1 & 20 & 6 & 0.192 & 0.197 & 0/6 \\
6 factors & D2 & 20 & 6 & 0.157 & 0.171 & 0/6 \\
6 factors & Random D2 & 20 & 6 & 0.214 & 0.216 & 0/120 \\
6 factors & GP-G & 20 & 6 & 0.330 & 0.334 & 0/6 \\
6 factors & GP-E & 20 & 6 & 0.248 & 0.252 & 0/6 \\
6 factors & Code D1 & 20 & 6 & 0.303 & 0.296 & 1/6 \\
6 factors & Code D2 & 20 & 6 & 0.389 & 0.384 & 1/6 \\
6 factors & Code GP-E & 20 & 6 & 0.462 & 0.468 & 1/6 \\
\midrule
6 factors & D1 & 32 & 6 & 0.298 & 0.301 & 0/6 \\
6 factors & D2 & 32 & 6 & 0.445 & 0.451 & 1/6 \\
6 factors & Random D2 & 32 & 6 & 0.465 & 0.466 & 17/120 \\
6 factors & GP-G & 32 & 6 & 0.548 & 0.547 & 1/6 \\
6 factors & GP-E & 32 & 6 & 0.537 & 0.538 & 1/6 \\
6 factors & Code D1 & 32 & 6 & 0.545 & 0.545 & 2/6 \\
6 factors & Code D2 & 32 & 6 & 0.650 & 0.650 & 3/6 \\
6 factors & Code GP-E & 32 & 6 & 0.697 & 0.697 & 2/6 \\
\bottomrule
\end{tabularx}%
}

\end{center}
At four factors and $B=8$, D2 recovery is 0.612 in utility units and 0.633 in
native units; GP-E gives 0.701 and 0.712. The paired code-equivalence gains
also persist under this transformation. These comparisons describe the effect
of utility normalization on the reported reconstructions.

\subsection{Signal and graph workflows}
\begin{figure}[!ht]
\centering
\includegraphics[width=\linewidth]{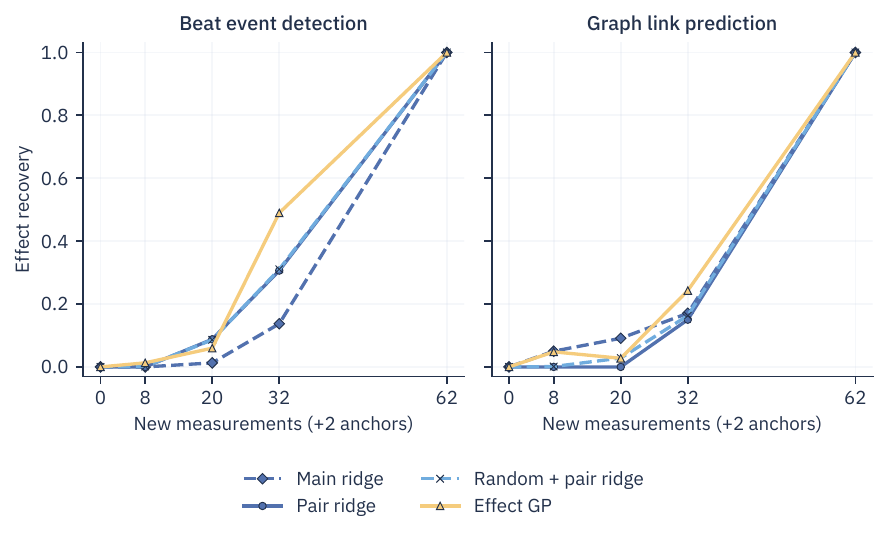}
\caption{Recovery across measurement budgets for beat detection and link
prediction. Each family contains four source instances, with complete native
references and the same experimental-design and reconstruction interfaces.}
\label{fig:additional}
\end{figure}

Figure~\ref{fig:additional} and the source profiles describe all eight added six-factor
cases. Order $\geq3$ variance is the uniform-cube Fourier share from
interactions of order three or higher~\citep{booleananalysis}. Curves average
seeds within source, then sources within family.
\begin{center}\small
\begin{tabularx}{\linewidth}{@{}lRRRR@{}}\toprule
\rowcolor{WWBHeader}
\textbf{Source} & \textbf{Min} & \textbf{Max} & \textbf{Sign reversals} & \textbf{Order $\geq3$ variance} \\
\midrule
mitdb\_100 & 0.502 & 1.000 & 4 & 0.162 \\
mitdb\_101 & 0.369 & 0.996 & 6 & 0.140 \\
mitdb\_102 & 0.554 & 1.000 & 5 & 0.386 \\
mitdb\_103 & 0.506 & 1.000 & 5 & 0.153 \\
graph\_lesmis & 0.500 & 0.903 & 6 & 0.419 \\
graph\_adjnoun & 0.500 & 0.747 & 5 & 0.084 \\
graph\_football & 0.500 & 0.892 & 5 & 0.265 \\
graph\_dolphins & 0.500 & 0.764 & 6 & 0.432 \\
\bottomrule\end{tabularx}

\end{center}

At $B=20$, two-family recovery is 0.0439 for pair ridge and 0.0435 for GP-E.
The complete controls cover all methods, budgets, and random seeds through
full enumeration.

\pagebreak
\subsection{Completed agent predictions on the added families}
\label{app:valid_added}
The frozen eight-source extension attempts four beat-detection and four graph
tasks at $B=32$. Six episodes deliver tables, covering three sources in each
family. The other sources are MIT-BIH record 103 and graph dolphins; they have
purchased observations and no submitted prediction table. The table
compares each submission with same-observation GP and source-matched D2,
grid-GP, and effect-GP controls.
\begin{center}
{\small
\begin{tabularx}{\linewidth}{@{}llRRRRR@{}}
\toprule
\rowcolor{WWBHeader}
\textbf{Family} & \textbf{Source} & \textbf{Agent $R$} & \textbf{Shared GP} & \textbf{D2} & \textbf{GP-G} & \textbf{GP-E} \\
\midrule
Beat detection & mitdb\_100 & 0.480 & 0.624 & 0.279 & 0.356 & 0.533 \\
Beat detection & mitdb\_101 & 0.515 & 0.642 & 0.336 & 0.509 & 0.536 \\
Beat detection & mitdb\_102 & 0.242 & 0.452 & 0.178 & 0.213 & 0.383 \\
\rowcolor{WWBStripe}
\multicolumn{2}{l}{\textbf{Beat detection mean}} & 0.412 & 0.573 & 0.264 & 0.359 & 0.484 \\
\midrule
Link prediction & graph\_lesmis & 0.000 & 0.185 & 0.000 & 0.467 & 0.000 \\
Link prediction & graph\_adjnoun & 0.528 & 0.596 & 0.425 & 0.591 & 0.565 \\
Link prediction & graph\_football & 0.053 & 0.234 & 0.163 & 0.348 & 0.406 \\
\rowcolor{WWBStripe}
\multicolumn{2}{l}{\textbf{Link prediction mean}} & 0.194 & 0.338 & 0.196 & 0.469 & 0.324 \\
\midrule
\rowcolor{WWBStripe}
\multicolumn{2}{l}{\textbf{Six-source macro}} & 0.303 & 0.455 & 0.230 & 0.414 & 0.404 \\
\bottomrule
\end{tabularx}
}

\end{center}

Shared GP improves all six delivered tables. Their family-macro recovery rises
from 0.303 to 0.455, with beat and graph gains of 0.160 and 0.145. The
all-attempt $R_{\rm all}$ values across four sources per family are 0.309 for
beat detection, 0.145 for graph prediction, and 0.227 with equal family weight.
The anonymous supplement retains the frozen plan and eight-attempt ledger.

\paragraph{Graph code-equivalence diagnostic.}
All four frozen graph references give identical native predictions and scores
for masks \texttt{000000} and \texttt{000001}. A CPU diagnostic applies this
relation to three completed $B=32$ submissions. Setting the second mask to
the measured first-mask score raises mean submitted recovery from 0.194 to
0.312; canonicalizing the pair for same-observation GP raises recovery from
0.338 to 0.516. Projection improves two submissions and GP all three. The
anonymous supplement contains the native and prediction audits.

\end{document}